\documentclass[conference]{IEEEtran}
\IEEEoverridecommandlockouts
\usepackage{cite}
\usepackage{amsmath,amssymb,amsfonts}
\usepackage{graphicx}
\usepackage{textcomp}
\usepackage[ruled,vlined]{algorithm2e}
\usepackage{subcaption}
\usepackage[normalem]{ulem}
\usepackage{multirow}
\usepackage{soul}
\newcommand{\ie}{\textit{i.e.}}
\newcommand{\eg}{\textit{e.g.}}
\usepackage{comment}
\usepackage{booktabs}
\newcommand{\CP}[1]{\ignorespaces}

\def\BibTeX{{\rm B\kern-.05em{\sc i\kern-.025em b}\kern-.08em
    T\kern-.1667em\lower.7ex\hbox{E}\kern-.125emX}}

\begin{document}

\title{Reducing Overconfidence Predictions in Autonomous Driving Perception\\
}

\author{\IEEEauthorblockN{~Gledson Melotti$^{1,2}$}
	\and
	\IEEEauthorblockN{~Cristiano~Premebida$^2$}
	\and
	\IEEEauthorblockN{~Jordan~J.~Bird$^3$}
	\and
	\IEEEauthorblockN{~Diego~R.~Faria$^4$}
	\and
	\IEEEauthorblockN{~Nuno Gon\c{c}alves$^5$}
	
	\thanks{$^1$ G.Melotti is with Federal Institute of Espirito Santo, Brazil. Email: \footnotesize gledson@ifes.edu.br}%
	
	\thanks{$^{2}$ C.Premebida is with the Institute of Systems and Robotics (ISR-UC), and the Dep. of Electrical and Computer Engineering at University of Coimbra-Portugal. Email: \footnotesize cpremebida@isr.uc.pt}%
	
	\thanks{$^{3}$ Computational Intelligence and Applications Research Group (CIA), Department of Computer Science, Nottingham Trent University, UK. Email: \footnotesize jordan.bird@ntu.ac.uk}%
	
	\thanks{$^{4}$ School of Physics, Engineering and Computer Science, University of Hertfordshire, UK. Email: \footnotesize d.faria@herts.ac.uk}%
	
	\thanks{$^{5}$ N.Gon\c{c}alves is with the Institute of Systems and Robotics (ISR-UC), Dep. of Electrical and Computer Engineering at University of Coimbra-Portugal, and Portuguese Mint and Official Printing Office, Lisbon-Portugal. Email: \footnotesize nunogon@isr.uc.pt}%
}

\maketitle

\begin{abstract}
In state-of-the-art deep learning for object recognition, Softmax and Sigmoid layers are most commonly employed as the predictor outputs. Such layers often produce overconfidence predictions rather than proper probabilistic scores, which can thus harm the decision-making of `critical' perception systems applied in autonomous driving and robotics. Given this, we propose a probabilistic approach based on distributions calculated out of the Logit layer scores of pre-trained networks which are then used to constitute new decision layers based on Maximum Likelihood (\textit{ML}) and Maximum a-Posteriori (\textit{MAP}) inference. We demonstrate that the hereafter called \textit{ML} and \textit{MAP} layers are more suitable for probabilistic interpretations than Softmax and Sigmoid-based predictions for object recognition. We explore distinct sensor modalities via RGB images and LiDARs (RV: range-view) data from the KITTI and Lyft Level-5 datasets, where our approach shows promising performance compared to the usual Softmax and Sigmoid layers, with the benefit of enabling interpretable probabilistic predictions. Another advantage of the approach introduced in this paper is that the so-called \textit{ML} and \textit{MAP} layers can be implemented in existing trained networks, that is, the approach benefits from the output of the Logit layer of pre-trained networks. Thus, there is no need to carry out a new training phase since the \textit{ML} and \textit{MAP}  layers are used in the test/prediction phase. The Classification results are presented using reliability diagrams, while detection results are illustrated using precision-recall curves.
\end{abstract}

\begin{IEEEkeywords}
Bayesian Inference; Confidence Calibration; Object Recognition; Perception System; Probability Prediction.
\end{IEEEkeywords}

\section{Introduction}
\label{sec:introduction}
\begin{figure}[!t]
	\centering
	\begin{subfigure}[b]{0.48\textwidth}
		\centering
		\includegraphics[width=\textwidth]{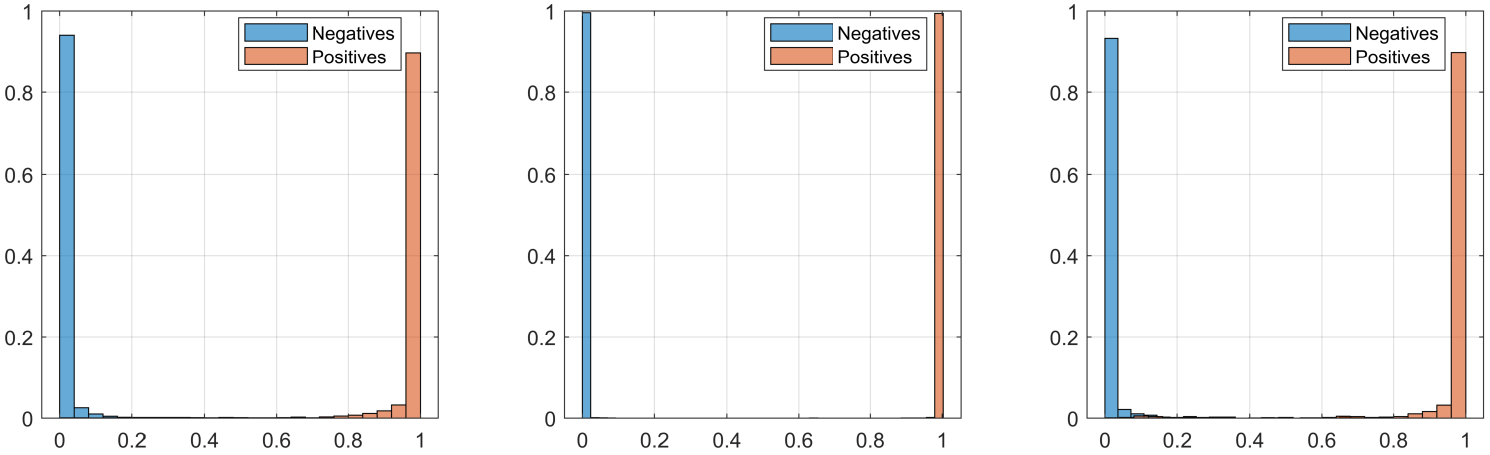}
		\caption{Softmax layer scores, RGB-KITTI.}
		\label{Softmax_RGB}
	\end{subfigure}
	\\
	\begin{subfigure}[b]{0.48\textwidth}
		\centering
		\includegraphics[width=\textwidth]{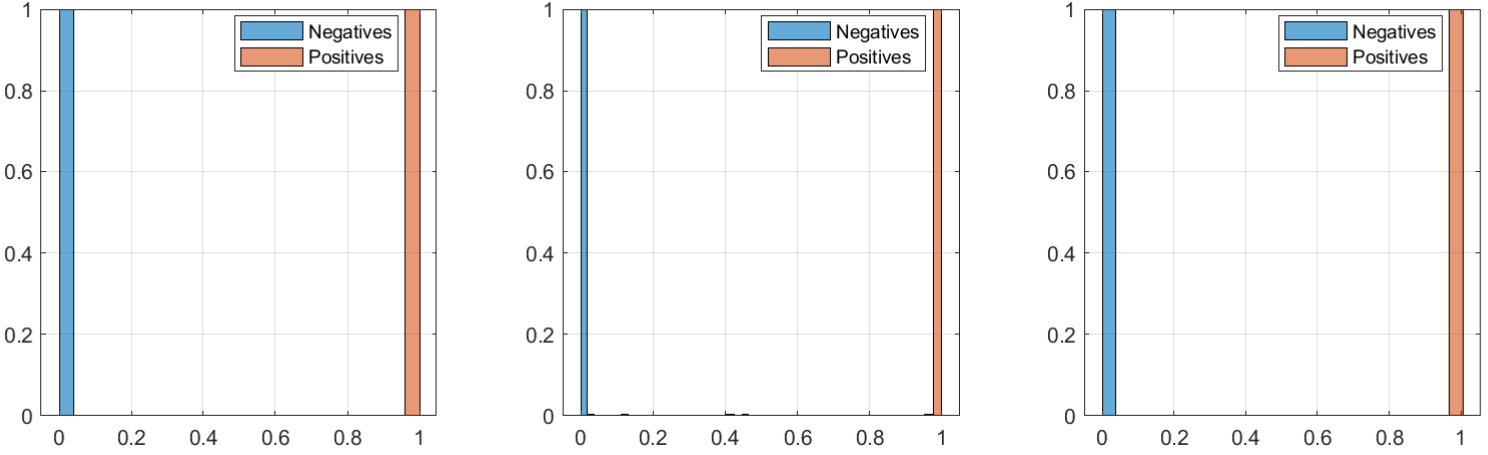}
		\caption{Softmax layer scores, RGB-LL5.}
		\label{Softmax_RGB_LL5}
	\end{subfigure}
	\caption{Graphs (a) and (b) are the Softmax prediction scores for the `pedestrian', `car' and `cyclist' classes (where the positives are in orange), showing evidence of overconfidence behavior. The bar-plots were obtained on a RGB image classification set from the KITTI and LL5 databases respectively.}
	\label{Softmax}
\end{figure}

Recent advances in deep learning and sensory technology (\eg, RGB cameras, LiDAR, radar, stereo, RGB-D, among others~\cite{Patel1,Bhatt,Janai2017,Shaoshan2017,Patel2}) have made remarkable contributions to perception systems applied to autonomous driving~\cite{hen,zwang,pcai,Schutera}. Perception systems include, but are not limited to, image and point cloud-based classification and detection~\cite{hpan,cai,pcai,zhouli,nie}, semantic segmentation~\cite{dfeng3,hen,CLI}, and tracking~\cite{zzuo,Santos}. Oftentimes, regardless of the type of network architecture or input modalities, most state-of-the-art CNN-based object recognition algorithms output normalized prediction scores via the Softmax layer~\cite{Su_2018_ECCV} \ie, the prediction values are in a range of $[0, 1]$, as shown in Fig. \ref{Softmax}. Furthermore, such algorithms are often implemented through deterministic neural networks, and the prediction itself does not consider the model's actual confidence for the predicted class in decision-making~\cite{Sensoy2018}. In fact, in most cases, the decision-making takes into account only the prediction value provided directly by a deep learning algorithm disregarding a proper level of confidence of the prediction (which is unavailable for most networks). Therefore, evaluating the prediction confidence or uncertainty is crucial in decision-making because an erroneous decision can lead to disaster, especially in autonomous driving where the safety of human lives are dependent on the automation algorithms.
 \noindent

Many works have pointed out Softmax layer overconfidence as an open issue in the field of deep learning~\cite{raudys2003reducing,bulatov2015reducing,kristiadi2020being,thulasidasan2019mixup}. Two main techniques have been suggested to mitigate the overconfidence in deep networks, calibration~\cite{gupta1,gupta2,oncalibration,Abdar,Mena,Bianca,Naeini} and regularization \cite{GabrielPereyra,Abdar,Mena}. Often, calibrations are defined as techniques that act directly on the resulting output of the network, while regularization are techniques that aims to penalize network weights through a variety of methods, which adds parameters or terms directly to the network cost/loss function~\cite{posch,zouyu2019,GabrielPereyra}. However, the paper proposed by~\cite{Gawlikowski} defines regularization techniques as a type of calibration. Consequently, the latter demands that the network must be retrained. 

\begin{figure}[!t]
	\centering
	\begin{subfigure}[b]{0.48\textwidth}
		\centering
		\includegraphics[width=\textwidth]{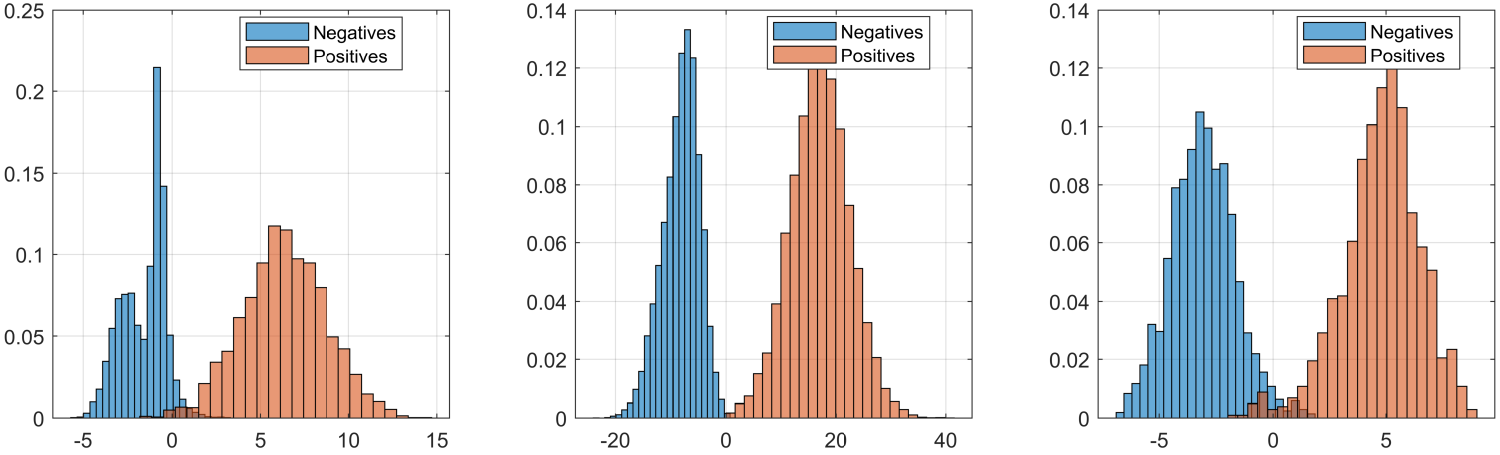}
		\caption{Logit layer scores, RGB-KITTI.}
		\label{HG_RGB}
	\end{subfigure}
	\\
	\begin{subfigure}[b]{0.48\textwidth}
		\centering
		\includegraphics[width=\textwidth]{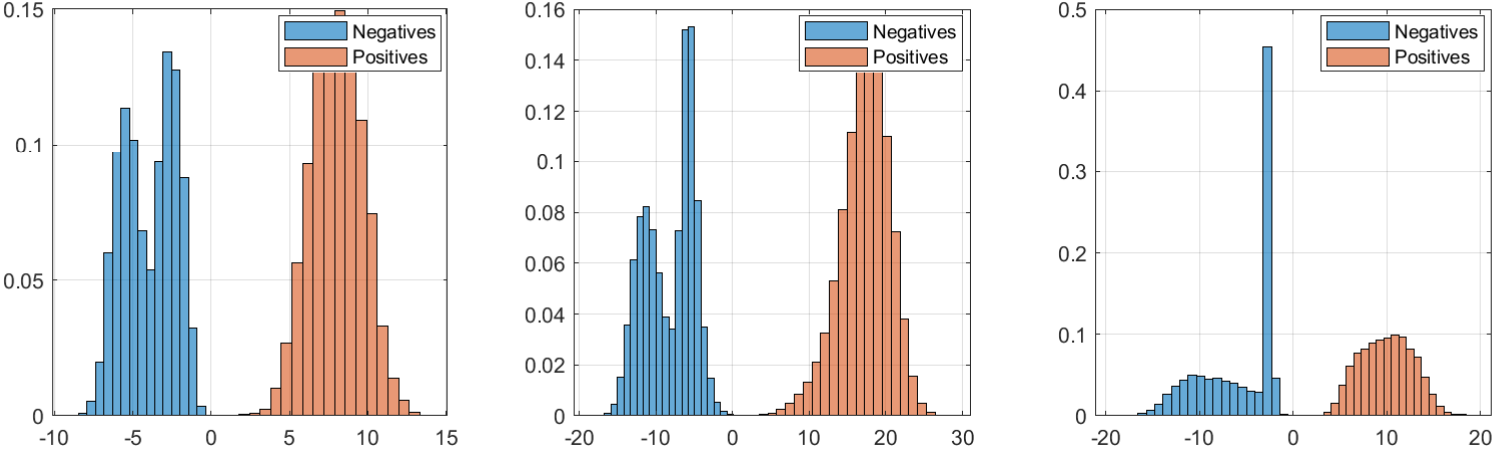}
		\caption{Logit layer scores, RGB-LL5.}
		\label{HG_RGB_LL5}
	\end{subfigure}
	\caption{Probability density functions (PDFs), using normalized histograms, for the Logit layers data on the training sets of the KITTI (a) and LL5 (b) datasets. The graphs are organized from left-right by classes (pedestrian, car and cyclist, where the positives are in orange) using the RGB modality.}
	\label{Sof_Log}
\end{figure}

The overconfidence problem is more evident in complex networks such as Convolutional Neural Networks (CNNs), particularly when using the Softmax layer as the prediction layer, thus generating ill-distributed outputs \ie, values close to either zero or one~\cite{oncalibration} which can be observed in Fig. \ref{Softmax_RGB} and Fig. \ref{Softmax_RGB_LL5}. We note that this is desirable when the true positives have higher scores. However, the counterpart problem is that `overconfidence networks' also generate high-score values for the objects erroneously detected or classified \ie, false positives. Given this problem, a question that arises, \textit{how can we guarantee prediction values that are `high' for true positives and, at the same time, `low' for false positives?} This question can be answered by analyzing the output of the network's Logit layer, which provides a smoother output than the Softmax layer. This can be observed within Figs. \ref{HG_RGB} and \ref{HG_RGB_LL5}. 

 \noindent

Following this, we can put a new question: \textit{although normalized outputs aim to guarantee a `probabilistic interpretation', how reliable are these predictions? Additionally, given an object belonging to a non-trained/unseen class (\eg, an unexpected object on the road), how confident is the model's prediction?} These are the key research questions explored in this work by considering the importance of having models grounded on interpretable probability assumptions to enable adequate interpretation of the outputs, ultimately leading to more reliable predictions and decisions. In terms of contributions, this paper introduces new prediction layers, designated Maximum Likelihood (\textit{ML}) and Maximum a-Posteriori (\textit{MAP})  layers, for deep neural networks, which provide a more adequate solution compared to state-of-the-art (Softmax or Sigmoid) prediction layers. Both \textit{ML} and \textit{MAP}  layers compute a single estimate, rather than a distribution. Moreover, this work contributes towards the advances of multi-sensor perception (RGB and LiDAR modalities) for autonomous perception systems~\cite{MartinICCV,geiger2012,melotti_icarsc} by proposing a probability-grounded solution that is practical in the sense it can be used in existing (\ie, pre-trained) state-of-the-art models such as Yolo~\cite{yolov420}.

It is important to emphasize that there is no need to retrain the neural networks when the approach described in this article is employed, because the \textit{ML} and \textit{MAP}  prediction layers produce outputs based on PDFs obtained from the Logits of already trained networks. Therefore, instead of using the traditional prediction layers (Softmax or Sigmoid) to predict the object scores on a test set, the \textit{ML} and \textit{MAP}  nonlinearities can be used to make the predictions for the objects scores. Thus, the proposed technique in this paper is practical given that a network has already been trained with Softmax (\textit{SM}) or Sigmoid (\textit{SG}) prediction layers. In other words, the \textit{ML} and \textit{MAP} layers depend on the Logit's outputs of the already trained network\footnote{A note for the reviewers: this paper is an extension of our workshop-paper \cite{gledson_eccv}, as well as an extension of the paper~\cite{GledsonMelotti}. The main difference between this paper and the two previously mentioned papers is in the analysis of the results through reliability diagrams, considering the expected calibration error, and maximum calibration error metrics. In addition, this paper considers a more detailed analysis regarding the predicted score values on out-of-training distribution test data (unseen class).}

In summary, the scientific contributions arising from this work are:
\begin{itemize}
	\item An investigation of the distribution of predicted values of the Logit and Softmax layers, for both calibrated and non-calibrated networks;
	\item An analysis of the predicted probabilities inferred by the proposed \textit{ML} and \textit{MAP} formulations, both for object classification and detection;
	\item An investigation of the predicted score values on out-of-training distribution test data (unseen/non-trained class);
	\item The proposed approach does not require the retraining of networks;
	\item Experimental validation of the proposed methodology through different modalities, RGB and Range-View (3D point clouds-LiDAR), for classification (using InceptionV3) and object detection (using YoloV4).
\end{itemize}

In this paper, we report on object recognition results showing that the Softmax and Sigmoid prediction layers do indeed sometimes induce erroneous decision-making, which can be critical in autonomous driving. This is particularly evident when `unseen' samples \ie, out-of-training distribution test data are presented to the network. On the other hand, the approach described here is able to mitigate such problems during the testing stage (prediction).

The rest of this article is structured as follows. The related work is presented in Section \ref{sec:Related_Work}, while the proposed methodology is developed in Section \ref{sec:method}. The experimental part and the results are reported in Section \ref{sec:experiments}, the conclusion is given in Section \ref{sec:conclusions}, while Section \ref{futurework} presents ideas to expand the proposed research, and finally Section \ref{Annex} (Appendix) presents results considering an extra experiment.

\section{Related Work}
\label{sec:Related_Work}

In this section, we review the key methodologies related to our proposed approach. We briefly discuss the uncertainties of neural networks based on the concepts of Bayesian inference, consequently defining the types of uncertainties that can be captured by the Bayesian Neural Networks (BNNs). Then, techniques for reducing overconfidence of prediction layers are presented as well, in particular the regularization and calibration techniques.

\subsection{Predictive Uncertainty}

Many deep learning methods used for perception systems (objects detection and recognition) do not capture the network uncertainties at training and test times. The Bayesian Neural Network (BNN) is an alternative to cope with uncertainties and it can be carried out through distinct approaches. One way is to obtain the posterior distribution using variational inference after defining a prior distribution to the network weights~\cite{posch,shridhar2019,Graves2011}. Another method is the ensemble of multiple networks with the same architecture and different training sets for estimating predictive uncertainty~\cite{Balaji2017}. 

Currently, many studies consider aleatory and epistemic uncertainties obtained through BNNs. Aleatory uncertainty is related to the inherent noise of observations (uncertainties arising from sensor inherent noise and associated with the distance of the object to be detected, as well as the object occlusion), while the epistemic ones explain the uncertainties in the model parameters (uncertainties of the model associated with the detection accuracy, showing the limitations of the model)~\cite{Kendall2017}. The formulation of aleatory and epistemic uncertainties with the aim of presenting confidence of predictions, which can capture the uncertainties in object recognition, can be done through BNNs, Shannon Entropy (uncertainty in the prediction output) and Mutual Information (confidence of the model in the output) to measure the uncertainty of the classification scores~\cite{Yarin2017,Feng2019,Feng2018}.

The uncertainty of a prediction can also be achieved through Monte Carlo dropout strategy, using the dropout layers at test time \ie, the predicted values depend on the randomly chosen connections between the neurons according to the dropout rate, that is, the same test example (an object) forwarded several times in the network can have different predicted values (the predicted values are not deterministic). In this way, it is possible to obtain the distribution, the average (final predicted value) and the variance (uncertainty)~\cite{yazo} for each example.

Differing from the aforementioned works, the approach proposed in this paper uses data obtained from the Logit layer of already trained/existing networks, to employ the concepts of Bayesian inference. The methodology proposed in this paper defines a final prediction value for each object and does not need to predict recurrently for the same object several times. Furthermore, the approach presented in the paper does not consider the distribution of the network weights, and thus, it is an efficient and practical approach. These advantages are clear when compared to traditional Bayesian neural networks and the Monte Carlo dropout strategy, because the novel strategy presented here avoids a high computational cost and at the same time does preserve the recognition/detection performance. Nevertheless, there are ongoing research on Bayesian neural networks that have reduced the computational cost through feature decomposition and memorization~\cite{jia}.

\subsection{Regularization and Calibration}

Another important component for the improvement of the predicted values are the regularization techniques that avoid overfitting and contribute to reduce overconfidence predictions, such as the transformation of network weights using $L1$ and $L2$~\cite{AndrewY} regularization, label and model regularization by a process of pseudo-label and self-training~\cite{zouyu2019}, label smoothing~\cite{lukasik20a}, knowledge distillation~\cite{geoffreyhinton}, architecture development where the network has to determine whether or not an example belongs to the training set, and specific cost mathematical formulation~\cite{corbi,LeaConf}. Other well-known regularization techniques are the Batch Normalization~\cite{BatchNormalization}, stochastic regularization techniques such as Dropout~\cite{drop}, multiplicative Gaussian noise~\cite{Srivastava}, and dropConnect~\cite{DropConnect}.

Alternatively, highly confident predictions can often be mitigated by calibration techniques such as temperature scaling ($TS$)~\cite{oncalibration}, by multiplying all the values of the logit vector by a scalar parameter, $\frac{1}{TS}>0$, for all classes, where the value of $TS$ is obtained by minimizing the negative log likelihood on the validation set; Isotonic Regression~\cite{isotonicregression} which combines binary probability estimates of multiple classes, thus jointly optimizing the bin boundary and bin predictions; Platt Scaling~\cite{plattscaling} which uses classifier predictions as features for a logistic regression model; Beta Calibration~\cite{betacalibration} which uses a parametric formulation that considers the Beta probability density function; compositional method (parametric and non-parametric approaches)~\cite{mix}, as well as the embedding complementary networks technique~\cite{Chen_2019_CVPR,LIANG2019}.

In this study, we reduce highly confident predictions on the test set by replacing the predicted values by Softmax and Sigmoid layers with the predicted values from \textit{ML} and \textit{MAP}  nonlinearities, obtaining a smoother score distribution for new objects. Such functions depend on the output of the network's Logit layer, by means of parametric (Gaussian functions) and nonparametric (normalized histograms) modeling. This is a post-training operation, that is, the novel inference functions proposed in this work do not modify the weights neither the cost function of the network and still provides very satisfactory results. This is an advantage over regularization techniques, since the \textit{ML} and \textit{MAP} layers do not require network retraining. The advantage of the approach proposed in this paper with respect to calibration techniques is to provide a smoother distribution of the predicted values without degrading the results.

\section{Proposed Method}
\label{sec:method}

This section presents the core of the proposed methodology \ie, the formulations for making predictions based on the novel \textit{ML} and \textit{MAP} prediction layers. The development of such a methodology begins with the concepts of probabilities, random variables, distribution function, probability density function and Bayes' theorem \ie, the background to develop the methodology proposed in this paper. In the second stage, we present the proposed method through formulations of the Maximum Likelihood (\textit{ML}) and Maximum a-Posteriori (\textit{MAP})  layers, as well as nonparametric and parametric mathematical modeling to define the posterior (likelihood-conditional) and prior probabilities. Finally, we present the network architectures, diagrams for evaluating the calibration of the proposed methodology, and the datasets that have been used in the experiments.

\subsection{A Brief Review of Probability and Density Functions}
\label{subsec.pdf}

The output scores $\mathbf{\textbf{x}}=$ $\{x_1,\ldots,x_{nc}\}$ of a supervised classification system with $nc$ classes, $\mathbf{\textbf{c}}=$ $\{c_1,\ldots,c_{nc}\}$ can be formulated according to a random experiment considering a sample space $\mathbf{S}$. The numerical outcome obtained from each element of $\mathbf{S}$ is related to a real number defined by the random variable (rv) $\mathbf{\textbf{x}}$ \ie, the output scores, which is conditioned to the rv $\bf{c}$. Formally, the rv is a function that maps each element of the sample space with a real number of the set $\mathbb{R}$, which can be simply expressed as $\mathbf{\textbf{x}}:\mathbf{S}\rightarrow\mathbb{R}$. In other words, an rv is a function $\mathbf{\textbf{x}}$ that outputs a real number $\mathbf{\textbf{x}}(\zeta )$ for each element $\zeta\in \mathbf{S}$ of a random experiment. From the sample space, an event (subset of $\mathbf{S}$) can be defined and associated with a probability $\mathbf{P}$ between the $[\xi$, $\xi+\Delta\xi]$ interval. Such probability is a distribution function and its derivative is the probability density function (PDF) $f_x(x=\xi)$, as in \eqref{pdf}~\cite{Papoulis}.
\begin{align}
	f_x(x=\xi)= \lim_{\Delta \xi \to 0} \cfrac{\mathbf{P}\{\xi \leq \mathbf{\textbf{x}} \leq \xi+\Delta \xi\}}{\Delta \xi},
	\label{pdf}
\end{align}
where $f_x(x=\xi)\geq 0$ $\forall$ $\xi$, considering $\xi$ continuous. The integral of \eqref{pdf} represents the probability $\mathbf{P}$ with the random variable $\mathbf{\textbf{x}}$ contained in the interval. Consequently, if the interval [$\xi$, $\xi+\Delta \xi$] is sufficiently small, the probability will be $\mathbf{P}\{\xi \leq \mathbf{\textbf{x}} \leq \xi+\Delta \xi\}\simeq f_x(x=\xi)\Delta \xi$ \ie, the probability of the random variable $\mathbf{\textbf{x}}$ is proportional to $f_x(x=\xi)$. Thus, the probability will be maximum if the interval [$\xi$, $\xi+\Delta \xi$] contains its value and $f_x(x=\xi)$ will be maximum. Such a value is the most likely value of $\mathbf{\textbf{x}}$.

Given the most likely value of the random variable $\mathbf{\textbf{x}}$, Maximum Likelihood (\textit{ML}) and Maximum a-Posteriori (\textit{MAP})  inferences can be obtained.
However, the random variable $\mathbf{\textbf{x}}$ is dependent of the variable $\mathbf{\textbf{c}}$ for the formulation of \textit{ML} and \textit{MAP}. Therefore, the density function is conditional to $\mathbf{\textbf{c}}$~\cite{Papoulis}, as formulated in \eqref{pdf2}:
\begin{align}
	f_x(x=\xi|\mathbf{\textbf{c}})=\lim_{\Delta \xi \to 0} \cfrac{\mathbf{P}\{\xi \le \mathbf{\textbf{x}} \le \xi+\Delta \xi|\mathbf{\textbf{c}}\}}{\Delta \xi}.
	\label{pdf2}
\end{align}

If the random variable is discrete, a probability mass function (PMF) is used instead of a probability density function (PDF). Assuming that the class conditional probability $P(\mathbf{\textbf{x}}|\mathbf{\textbf{c}})$ (likelihood) and the prior are known, the posterior probability $P(\mathbf{\textbf{c}}|\mathbf{\textbf{x}})$ can be obtained through Bayes' rule
\begin{align}
	P(\mathbf{\textbf{c}}|\mathbf{\textbf{x}})=\cfrac{P(\mathbf{\textbf{x}}|\mathbf{\textbf{c}})P(\mathbf{\textbf{c}})}{P(\mathbf{\textbf{x}})},
	\label{bayes0}
\end{align}
where $P(\mathbf{\textbf{c}})$ is the prior probability, $P(\mathbf{\textbf{x}})\neq 0$ is the marginal probability defined by $\int P(\mathbf{\textbf{x}}|\mathbf{\textbf{c}})(\mathbf{\textbf{c}})d\textbf{c}$, that often can be determined by law of the total probability~\cite{bishop}. Thus, \eqref{bayes0} can be re-written using the \textit{per-class} expression:
\begin{align}
	P(c_i|\mathbf{\textbf{x}}) = \cfrac{P(\mathbf{\textbf{x}}|c_i)P(c_i)}{{\sum\limits_{i=1}^{nc}P(\mathbf{\textbf{x}}|c_i)P(c_i)}}.
	\label{bayes1}
\end{align}

In this work, the goal is to use \eqref{bayes1} to make inferences on the test set about the `unknown' rv $\mathbf{\textbf{c}}$ from the dependence with $\mathbf{\textbf{x}}$ \ie, the value of the posterior distribution of $\mathbf{\textbf{c}}$ is determined after observing the value of $\mathbf{\textbf{x}}$.

% % New Subsec name

\subsection{ML and MAP Prediction Layers}

\begin{figure}[!t]
	\centering
	\begin{subfigure}[b]{0.48\textwidth}
		\centering
		\includegraphics[width=\textwidth]{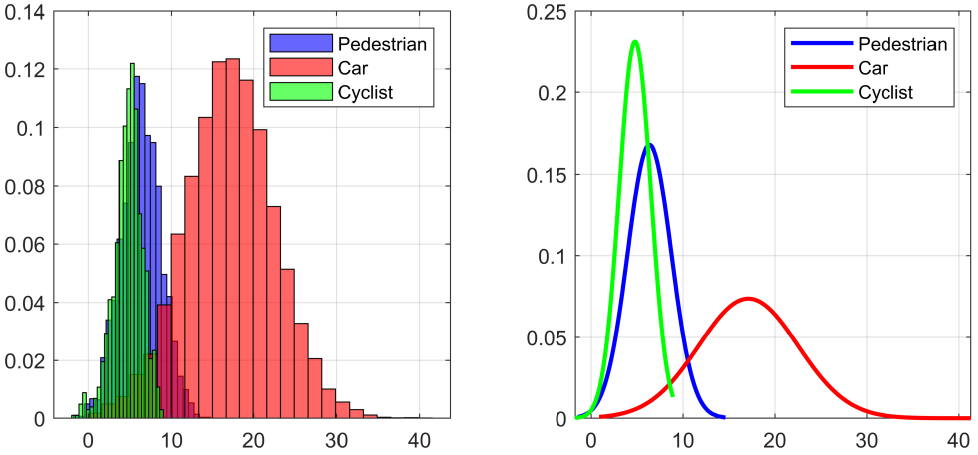}
		\caption{Logit layer output: RGB-KITTI.}
		\label{HG_PDF_RGB}
	\end{subfigure}
	\hfill
	\begin{subfigure}[b]{0.48\textwidth}
		\centering
		\includegraphics[width=\textwidth]{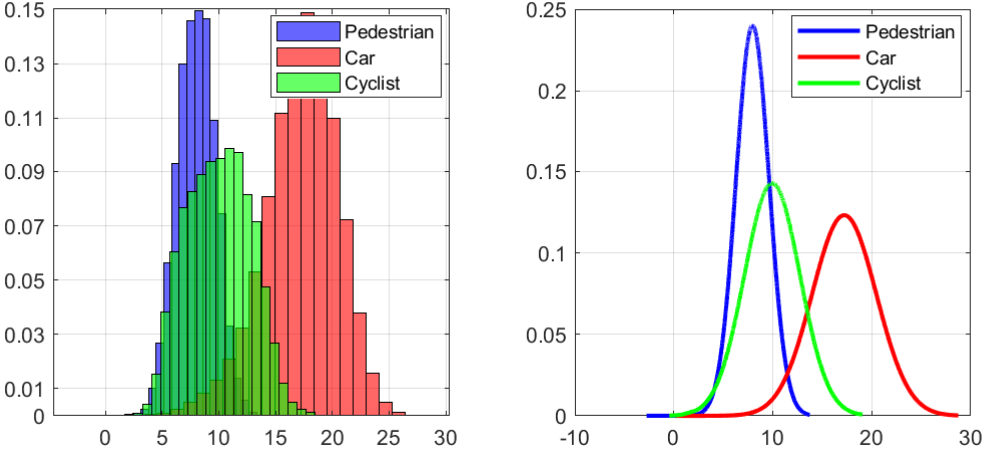}
		\caption{Logit layer output: RGB-LL5.}
		\label{HG_PDF_DM}
	\end{subfigure}
	\caption{From left-right respectively, normalized histogram-based densities and Gaussian densities calculated on the Logit layer values, for each class, on the training set (here for the RGB modality). On the $1^{st}$ row, we have the densities on the KITTI set while the $2^{nd}$ row shows the densities on the LL5 training set.}
	\label{HG_PDF}
\end{figure} \noindent

The proposed \textit{ML} and \textit{MAP}  layers make inference based on PDFs obtained from the Logit layer prediction scores by using the training set. This is illustrated in Fig. \ref{HG_PDF}, where the horizontal axes represent the random variable $\mathbf{\textbf{x}}$ and the vertical axes are the normalized frequency of the amount of objects in the classification and detection datasets. We can observed that the distribution scores from the Logit layer are far more appropriate to represent a PDF (as shown in Fig. \ref{Sof_Log}). Therefore, the \textit{ML} and \textit{MAP} layers are more adequate to perform probabilistic inference in regard to permitting decision-making under uncertainty, which is particularly relevant in autonomous driving and robotic perception systems.

As noted in \eqref{bayes1}, the posterior probability depends on the class conditional probability (likelihood function) and on the prior probability \ie, the \textit{MAP} estimated depends on a distribution for both the likelihood and prior, while \textit{ML} only depends on $P(\mathbf{\textbf{x}|\textbf{c}})$, because $P(\mathbf{\textbf{c}})$ is usually assumed to be uniform and identically distributed. The probabilities $P(\mathbf{\textbf{x}}|\mathbf{\textbf{c}})$ are modeled by means of non-parametric estimates over the predicted scores of the Logit layer for each class, as showed in the first column of Fig. \ref{HG_PDF}. These estimates are obtained on the training set, through normalized histograms (\ie, discrete densities defined by a single parameter - the number of bins) for each modality, as shown in the Table \ref{NB_AS}. 

Histograms are graphical ways of summarizing or describing a variable in a simple way, in other words, histograms show how variables (in this case, the network logits) are distributed, revealing modes and bumps, as well as information about the frequencies of observations. As said by C. Bishop~\cite{bishop}, `we can view the histogram as a simple way to model a probability distribution given only a finite number of points drawn from that distribution'. Often, the bins of a histogram are chosen to have the same width thus, the only (single) parameter left is the number of bins (nbins). To do so, nbins can be mathematically determined by means of the mean squared error (MSE-expected value of the squared error)~\cite{ScottDavidW}. However, for our methodology, we have chosen nbins empirically to guarantee a result very close to or better than the results provided by the \textit{SM} and \textit{SG} layers and, in addition, to generate smoother distribution by adding the parameter $\lambda$. Thus, the process of estimating the number of bins and $\lambda$ (the additive smoothing factor) have been defined empirically by verifying which combinations would not degrade the results. So, these two parameters were defined empirically for each dataset/modality, as well as for each of the \textit{ML} and \textit{MAP} layers. 
\begin{table}[!t]
	\begin{center}
		\caption{Number of bins and smoothing parameter ($\lambda$) for \textit{ML} and \textit{MAP}  layers.}
		\label{NB_AS}
		\begin{scriptsize}
			\begin{tabular}{ccccc}
				\toprule
				\multicolumn{5}{c}{Maximum Likelihood} \\ \hline
				& RGB Modalitiy &  & RV Modality &  \\ \hline
				Dataset & Bins & \begin{tabular}[c]{@{}c@{}}Additive\\ Smoothing\end{tabular} & Bins & \begin{tabular}[c]{@{}c@{}}Additive\\ Smoothing\end{tabular} \\ \hline
				KITTI & $25$ & $1.0\times 10^{-2}$ & $25$ & $1.0\times 10^{-2}$ \\ \hline
				LL5 & $25$ & $1.0\times 10^{-2}$ & $30$ & $1.0\times 10^{-2}$ \\ \hline
				\multicolumn{5}{c}{Maximum a-Posteriori} \\ \hline
				& RGB Modalitiy &  & RV Modality &  \\ \hline
				Dataset & Bins & \begin{tabular}[c]{@{}c@{}}Additive\\ Smoothing\end{tabular} & Bins & \begin{tabular}[c]{@{}c@{}}Additive\\ Smoothing\end{tabular} \\ \hline
				KITTI & $25$ & $1.0\times 10^{-2}$ & $25$ & $1.0\times 10^{-2}$ \\ \hline
				LL5 & $25$ & $1.0\times 10^{-2}$ & $30$ & $1.0\times 10^{-2}$ \\
				\bottomrule
			\end{tabular}
		\end{scriptsize}\noindent
	\end{center}
\end{table}

\begin{figure}[!t]
	\centering
	\includegraphics[scale=0.695]{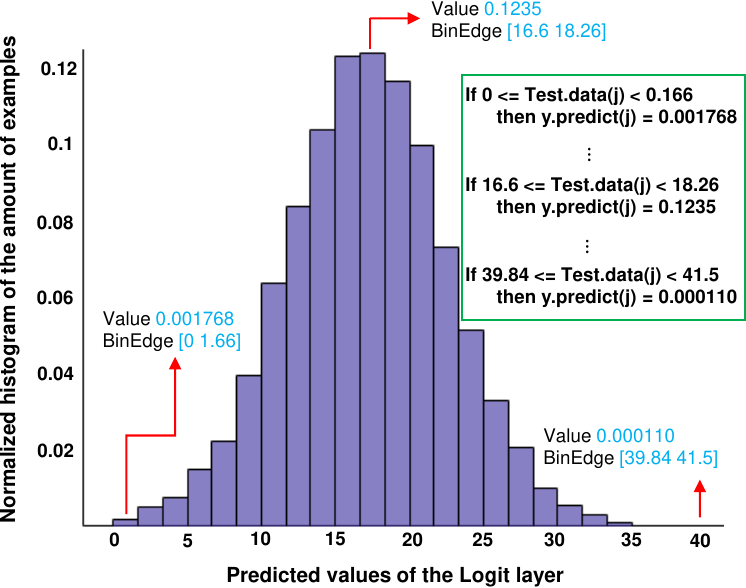}
	\caption{Obtaining probability values of a normalized histogram generated with the training data of the Logit layer.}
	\label{Prior}
\end{figure}

Each predicted value on the test set from the Logit layer has a score value corresponding to its bin range in the respective class histogram, which is illustrated in Fig. \ref{Prior}. For the \textit{MAP} layer, the prior is modeled by a Gaussian distribution that guarantees a smoother distribution of the prediction values, as observed within the second column of Fig. \ref{HG_PDF}. Thus, $P(\mathbf{\textbf{c}}) \sim \mathcal{N}(\mathbf{\textbf{x}}|\mu,\,\sigma^{2})$ with mean $\mu$ and variance $\sigma^{2}$ is calculated per class, from the training set. The modeling with different distribution techniques, Gaussian distribution and normalized histogram, aims to capture complementary information from the training data, where the maximum values per classes in the normalized histograms are different from the maximum values of the Gaussian distributions (Fig. \ref{HG_PDF}).

The normal distribution is feasible for modeling an unknown distribution because it has a maximum entropy. Thus, the greater entropy can guarantee a more informative distribution and at the same time less confident information around the mean, that is, it contributes to the reduction of the overconfidence inferences. Defining otherwise, the events most likely to happen have low information content \ie, low entropy. Therefore, a Gaussian distribution was defined for prior $P(c_i)$ to express a high degree of uncertainty\footnote{The amount of uncertainty can be quantified, for example, using Shannon's entropy for a probability distribution.} in the value of variable $\mathbf{\textbf{c}}$ before observing the data. Furthermore, a prior distribution with high entropy is said to be a prior distribution with high variance~\cite{bishop}.
 \noindent

Additionally, to avoid the `zero' probability problem, as well as to incorporate some uncertainty level in the final prediction, the Additive Smoothing method ($\lambda$)~\cite{AdditiveS, SmoTec, Lidstone} (also defined as Laplace smoothing) is implemented during the \textit{ML} and \textit{MAP} predictions. The values assigned for the Additive Smoothing are shown in Table \ref{NB_AS}, does not depend on previous information of the training set. This value was determined empirically \ie, by observing which value would preserve approximately the `original' distribution without compromising the final result. The probability estimates with the Additive Smoothing are shown in \eqref{bayes2} and \eqref{bayes3}, \ie, a small correction is incorporated into the \textit{ML} and \textit{MAP} estimate. Consequently, no prediction will have a `zero' probability, no matter how unlikely.

\textit{ML} layer is straightforwardly calculated by normalizing $P(\mathbf{\textbf{x}}|\mathbf{\textbf{c}})$ by the $P(\mathbf{\textbf{x}})$ during the prediction phase, as in \eqref{bayes2}, since the priors $P(\mathbf{\textbf{c}})$ are set uniformly and identically distributed for the set of classes $\mathbf{\textbf{c}}$,
\begin{align}
	ML = arg \max_{i} \cfrac{(P(\mathbf{\textbf{x}}|c_i)+\lambda)}{\sum\limits_{i=1}^{nc}(P(\mathbf{\textbf{x}}|c_i)+\lambda)}.
	\label{bayes2}
\end{align}

Alternatively, the inference using \textit{MAP} layer is given in \eqref{bayes3} as follows,
\begin{align}
	MAP = arg \max_{i} \cfrac{(P(\mathbf{\textbf{x}}|c_i)P(c_i)+\lambda)}{\sum\limits_{i=1}^{nc}(P(\mathbf{\textbf{x}}|c_i)P(c_i)+\lambda)}
	\label{bayes3}.
\end{align}

\begin{algorithm}
	\small
	\SetAlgoNoEnd
	\caption{compute \textit{ML} and \textit{MAP}.}
	\textbf{Input}
	\begin{itemize}
		\item Number of classes used in training ($nc$);
		\item Number of histogram bins (nbins);
		\item Values of the Logit layer on the training set ($train_{Lg}$);
		\item PDF's parameters (normalized histogram and normal on the training set, see Fig. \ref{HG_PDF});
		\item Values of the Logit layer on the testing set ($test_{Lg}$).
		\item Additive smoothing ($\lambda$).
	\end{itemize}
	\textbf{Output}
	\begin{itemize}
		\item Maximum Likelihood (\textit{ML}) and Maximum a-Posteriori (\textit{MAP}).
	\end{itemize}
	\textbf{Getting the normalized frequency histograms:\\
		$hc \gets histogram(ScoresLogitsTrain(classes))$;\\
		\textbf{Getting the edge values of each bin of each histogram:}\\
		$BinLow \gets BinEdgesLow(hc)$;\\
		$BinHigh \gets BinEdgesHigh(hc)$;\\
		\textbf{Getting the normalized frequency values of each bin of each histogram:}\\
		$Values \gets Values(hc)$};\\
	\textbf{Getting the likelihood:}\\
	$P(\mathbf{x|C}) \gets zeros(size(test_{Lg}),nc)$;\\
	$Y \gets ScoresLogitsTest$;\\
	\For{$k \gets 1:size(test_{Lg})$}{
		\For{$cla \gets 1:nc$}{
			\For{$i \gets 1:size(Values)$}{
				\If{$(BinLow(cla,i) \leqslant Y(k,cla)) \, \& \, (Y(k,cla) < BinHigh(cla,i))$}
				{
					$P(\mathbf{x}|C)(k,cla) \gets Values(cla,i)$\;
				}
				\textbf{end}
			}
			\textbf{end}
		}
		\textbf{end}
	}
	\textbf{end}\\
	\textbf{Getting the Prior:}\\
	$P(\mathbf{C}) \gets \mathcal{N}(test_{Lg}|[\mu_{train},\,\sigma^{2}_{train}])$;\\
	\textbf{Calculating the \textit{ML} and \textit{MAP}:}\\
	%\textbf{Additive smoothing}\\
	$ML \gets P(\mathbf{x|C}) + \lambda$;\\ 
	$ML \gets (ML/\mbox{sum}(ML))$;\\
	$MAP \gets P(\mathbf{x|C})P(\mathbf{C}) + \lambda$;\\ 
	$MAP \gets (MAP/\mbox{sum}(MAP))$;\\
	\label{alg1}
\end{algorithm}

The sequential steps for calculating the \textit{ML} and \textit{MAP} is summarized within Algorithm \ref{alg1}, where class-conditional $P(\mathbf{\textbf{x}}|\mathbf{\textbf{c}})$ is modelled by a normalized histogram. On the other hand, to get the maximum posterior probabilities (\textit{MAP}) the priors are modelled by normals $\mathcal{N}(test_{Lg}|\mu_{train},\,\sigma^{2}_{train})$, where the sub-index $Lg$ indicates that the data is obtained from the Logit layer (layer before the network prediction layer). Both the likelihood and prior are extracted from the Logit layer using the training data\footnote{The code for training the network, obtaining the logit layers and computing the \textit{ML} and \textit{MAP} layers are available at github.com/gledsonmelotti/ML-MAP-Layers-for-Probabilistic.}.

\subsection{CNN Architectures for Object Recognition}

\begin{figure*}[!t]
	\centering
	\includegraphics[scale=0.80]{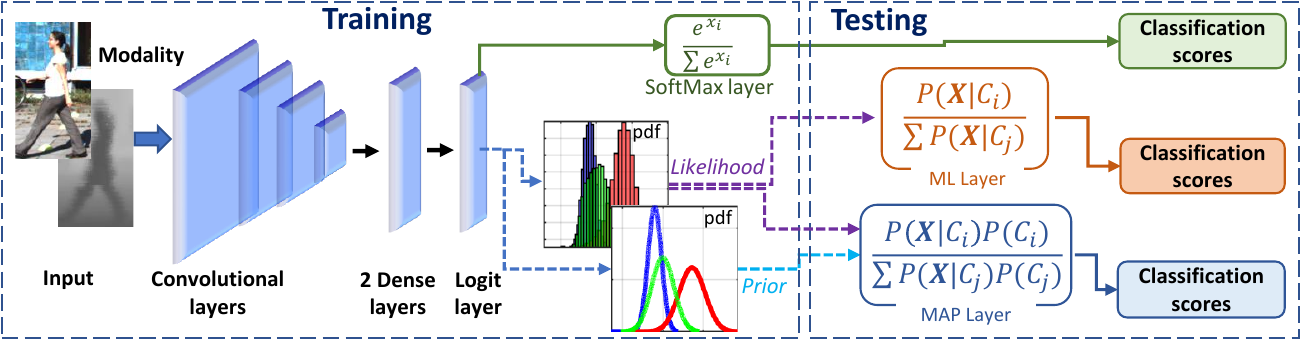}
	\caption{Inception V3 CNN representation with Logit and Softmax layers, Maximum Likelihood (\textit{ML}) and  Maximum a-Posteriori (\textit{MAP})  layers. CNN's training was done with the Softmax layer. After training, the Softmax layer was replaced by the \textit{ML} and \textit{MAP}  \ie, the CNN was not trained with the \textit{ML} and \textit{MAP}  layers.}
	\label{Inception V3}
\end{figure*} \noindent

Experiments in~\cite{oncalibration} suggested that the greater the number of layers and neurons, the more overconfidence the result will be. However, the experiments that we have conducted show that even when reducing the amount of neurons and filters in the dense and convolutional layers, the network can still produce overconfidence in the predicted values, as can be observed in Fig. \ref{Softmax}. This conclusion was reached by training the Inception V3 CNN~\cite{incv3} and reducing the number of filters and neurons/units. Regarding object detection, the model Yolo V4~\cite{yolov420} was trained to detect cars, cyclists, and pedestrians, with predictions based on the SG layer.

The experiments reported throughout the remainder of this work were based on the premise that, after training the network, the proposed \textit{ML} and \textit{MAP}  layers then replace the \textit{SM} and \textit{SG} prediction layers on the test set, only, according to Fig. \ref{Inception V3}.

\subsection{Reliability Diagram}
\begin{figure}[!t]
	\centering
	\includegraphics[scale=0.43]{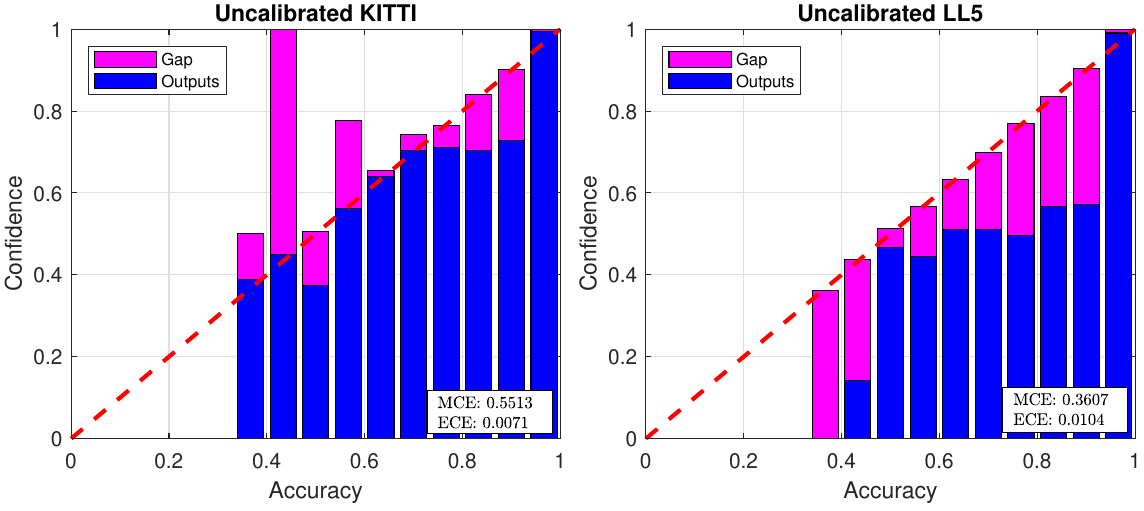}
	\caption{Reliability diagrams for the RGB modality on the testing set using the Softmax layer (\textit{SM}). On the left, uncalibrated model for KITTI dataset, and on the right uncalibrated model for LL5.}
	\label{UncRD}
\end{figure} \noindent

Typically, post-calibration predictions are analyzed in the form of reliability diagram representations~\cite{beytscal,oncalibration}, which illustrate the relationship of the model's prediction scores in regard to the true correctness likelihood~\cite{Niculescu}, as shown in Fig. \ref{UncRD}. Reliability diagrams show the expected accuracy of the samples as a function of confidence \ie, the maximum value of the prediction function.

The scores (predicted values) are grouped into $M$ bins (histogram) in the reliability diagrams. Each sample (classification score of an object) is allocated within a bin, according to the maximum prediction value (prediction confidence). Each bin has a range $I_m=\big (\frac{(m-1)}{M},\frac{m}{M}\big ]$, where $m={1,..,M}$. The accuracy is calculated in each range $I_m$, as well as the average confidence $conf_{average}=\frac{1}{BM}\sum_i\hat{p_i}$, where $\hat{p_i}$ is the confidence for sample $i$ and $BM$ is the amount of objects in each $I_m$. In addition, a gap can be obtained \ie, the difference between accuracy and average confidence in each range ($I_m$). Thus, the greater the gap, the worse the calibration result in the respective bin. Furthermore, through reliability diagrams, it is possible to obtain calibration errors, such as the Expected Calibration Error (ECE) and the Maximum Calibration Error (MCE):
\begin{align}
	\label{ece}
	ECE=\sum\limits_{m=1}^{M}\cfrac{|BM|}{n}|acc(BM)-conf(BM)|,\\
	\label{mce}
	MCE=\max_{m\in\{1,...,M\}}|acc(BM)-conf(BM)|,
\end{align}
where n is the number of samples.

Moreover, the reliability diagrams illustrate the identity function (diagonal-dashed line) that represents a perfectly calibrated output, while any deviation from the diagonal represents a calibration error~\cite{beytscal,oncalibration}.%, as shown in Fig. \ref{UncRD} 
%and \ref{UC_TS_HG_RGB_LL5}. The $1^{st}$ row, from left to right, of Fig. \ref{UC_TS_HG} show the uncalibrated ($UC$) network scores.
%, while the center are the calibrated ones. The last column within Fig. \ref{UC_TS_HG} shows the distribution of scores on the test set, which are still overconfident even after TS calibration has been applied. Consequently, such calibration does not guarantee a good balance of the prediction scores and may jeopardize adequate probability interpretation.

\subsection{Benchmarking Datasets}
\label{subsec:Benchmarking Datasets}
\begin{table}[!t]
	\begin{center}
		\caption{KITTI and LL5 dataset for classification: number of objects per class and subsets.}
		\label{dataset}
		\begin{scriptsize}
			\begin{tabular}{cccc}
				\toprule
				\multicolumn{4}{c}{ \textbf{KITTI dataset - 7481 Frames} } \\
				%\hline \hline
				& \textbf{Car} & \textbf{Cyclist} & \textbf{Pedestrian} \\ \hline
				\textbf{Training}    & $18103$ & $1025$  & $2827$ \\
				\textbf{Validation}  & $2010$  & $114$   & $314$ \\
				\textbf{Testing}     & $8620$  & $488$   & $1346$ \\
				
				& \multicolumn{3}{c}{\textbf{Non-trained (`unseen-adversarial') objects}} \\
				%\hline
				& \textbf{Tram/Truck/Van} & \textbf{Tree/lamppost} & \textbf{Person-sitting} \\
				\hline
				\textbf{Training}    & - & -  & - \\
				\textbf{Validation}  & -  & -   & - \\
				\textbf{Testing}     & $511/1094/2914$  & $45$   & $222$ \\
				\toprule
				\multicolumn{4}{c}{ \textbf{LL5 dataset - 158757 Frames} } \\
				%\hline \hline
				& \textbf{Car} & \textbf{Cyclist} & \textbf{Pedestrian} \\ \hline
				\textbf{Training}    & $208501$ & $7199$  & $9031$ \\
				\textbf{Validation}  & $23167$  & $800$   & $1003$ \\
				\textbf{Testing}     & $193012$ & $9238$  & $9199$ \\
				& \multicolumn{3}{c}{\textbf{Non-trained (`unseen-adversarial') objects}} \\
				%\hline
				& \textbf{Bus/OtherVehicle/Truck} & \textbf{Tree/lamppost} & \textbf{Motorcycle} \\
				\hline
				\textbf{Training}    & - & -  & - \\
				\textbf{Validation}  & -  & -   & - \\
				\textbf{Testing}    & $5257/2785/10890$  & $45$   & $217$ \\
				\bottomrule
			\end{tabular}
		\end{scriptsize}\noindent
	\end{center}
\end{table}
\noindent
\raggedbottom

\begin{figure}[!t]
	\centering
	\begin{subfigure}[b]{0.48\textwidth}
		\centering
		\includegraphics[width=\textwidth]{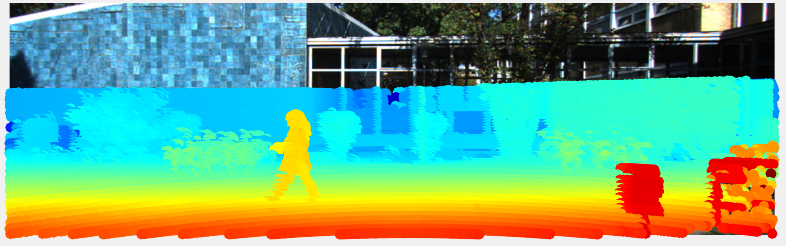}
		\caption{$3D$ point cloud projected on the $2D$ image plane.}
		\label{frame58_RGB}
	\end{subfigure}
	\\
	\begin{subfigure}[b]{0.48\textwidth}
		\centering
		\includegraphics[width=\textwidth]{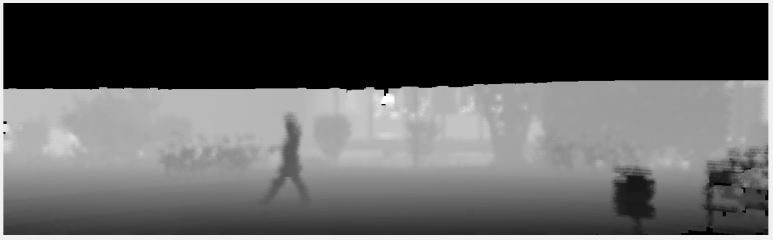}
		\caption{Range-view (RV) after upsampling the point cloud.}
		\label{frame58_DM}
	\end{subfigure}
	\hfill
	\caption{Example from the KITTI dataset. Representations of a `raw' point-cloud (a) in image coordinates and the upsampled range-view (b) obtained using the bilateral filter.}
	\label{frame58}
\end{figure} \noindent

\begin{figure}[!htbp]
	\centering
	\begin{subfigure}[b]{0.475\textwidth}
		\centering
		\includegraphics[width=\textwidth]{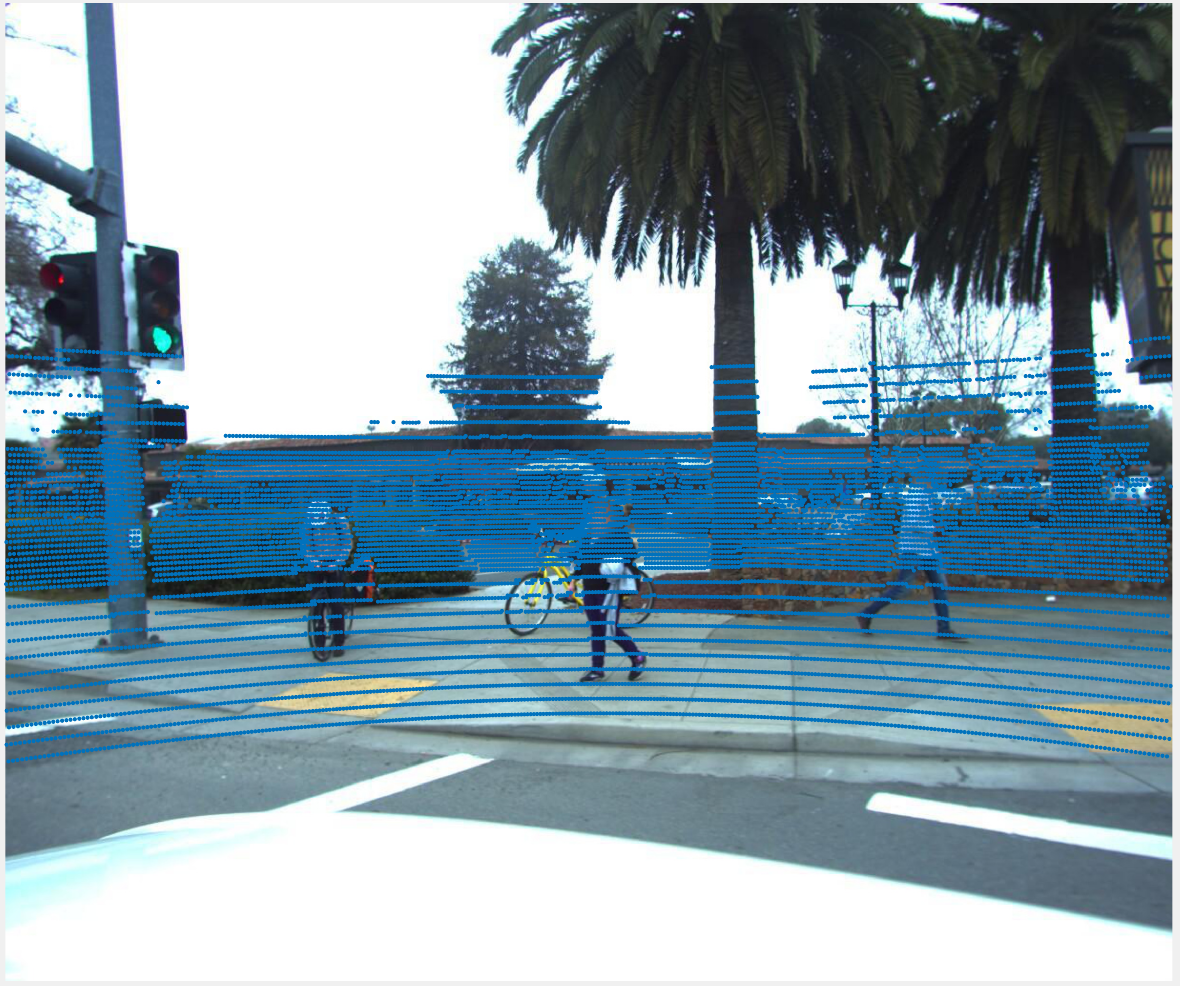}
		\caption{$3D$ point cloud projected on the $2D$ image plane.}
		\label{CamFront}
	\end{subfigure}
	\\
	\begin{subfigure}[b]{0.48\textwidth}
		\centering
		\includegraphics[width=\textwidth]{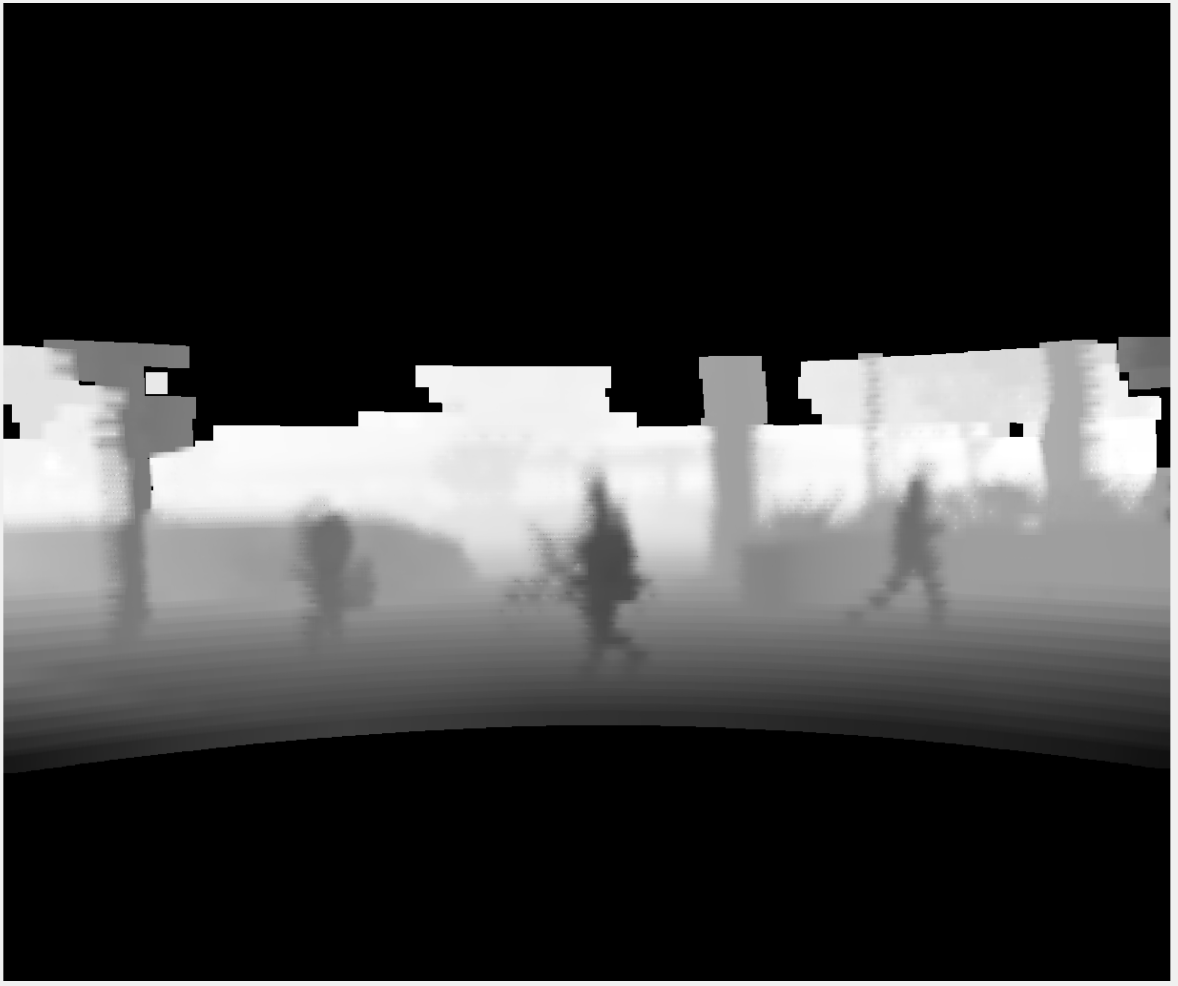}
		\caption{Range-view (RV) for a 40-channels LiDAR.}
		\label{CamFrontDepth}
	\end{subfigure}
	\hfill
	\caption{Example from the LL5 dataset. In (a) the $3D$ point clouds are in pixel-coordinates, and (b) shows the respective range-view after applying the bilateral filter.}
	\label{LL5}
\end{figure} \noindent

A key contribution to the growing improvement of perception systems for autonomous driving is the availability of representative datasets of different modalities, such as RGB, LiDAR, and radar~\cite{nuscenes20,astar3d,Song_2019_CVPR,Yang_2019_CVPR,H3D2019,Mapillary2017}. In this work, we used the KITTI Vision Benchmark Suite-$2D$ object~\cite{geiger2012} and Lyft Level-5 (LL5) Perception~\cite{lyftlevel5,lyft2019} datasets. The classes of interest were pedestrians, cars, and cyclists. Table \ref{dataset} shows the number of objects cropped from both the RGB and range-view (depth from the LiDAR modality) images. In addition, some extra objects from the unseen/non-trained classes (not used during training), such as a person sitting, tram, truck, van, tree, lamppost, signpost, bus, and motorcycle classes were classified in the test/prediction phase, to verify the erroneous overconfidence from the prediction layers of the trained networks. Such a class can be understood as an `adversarial' class; \textit{Note that this research did not carry out any study involving adversarial network architectures.}

Range-view images were obtained by a coordinate transformation of the $3D$ point clouds on the $2D$ image plane followed by an upsample of the projected points. The upsample was performed using a bilateral filter, and considered a mask size $13 \times 13$ (sliding-windows)~\cite{melotti_icarsc,me1,me2,cp_high} for t
he KITTI dataset and a mask size $23 \times 23$ for LL5 dataset. Examples of these operations can be observed in Fig. \ref{frame58} and Fig. \ref{LL5}, respectively.

As a way to validate the proposed methodology for object detection, the KITTI Vision Benchmark Suite-$2D$ object was used. The respective dataset was divided into $3367$ frames for the training dataset, $375$ frames for the validation dataset and $3739$ frames for the test dataset.

\section{Evaluation and Results}
\label{sec:experiments}

The output scores of the CNN indicate a degree of certainty of the given prediction. The level of certainty can be defined as the confidence of the model, and in an object recognition problem, represents the maximum value within the prediction layer. However, the output scores may not always represent a reliable indication of certainty with regard to a given class, especially when unseen (non-trained) objects occur in the prediction stage; this is particularly relevant for a real world application involving autonomous robots and vehicles, since unpredictable objects are likely to be encountered which would be misclassified by prediction layers with a high degree of certainty. With this in mind, in addition to the trained classes (pedestrian, car, and cyclist), a set of unseen objects were introduced into the classification dataset, according to Subsection \ref{subsec:Benchmarking Datasets}. Regarding the object detection, the unseen classes are already contained in the dataset's own frames. Unlike the results reported on the classification dataset, the object detection results are presented by means of precision-recall curves considering the easy, moderate, and hard cases, according to the devkit-tool provided by the KITTI benchmark.

\subsection{Results on Object Classification}
\begin{table*}[!t]
	\begin{center}
		\caption{Comparison between the classifications obtained by the \textit{SM} layer, \textit{ML} and \textit{MAP} layers in terms of average F-score and $FPR$ ($\%$). The performance measures on the `unseen' dataset are the average and the variance of the prediction scores.}
		\begin{tabular}{l c c c c c c}
			\toprule
			\multicolumn{7}{c}{ \textbf{KITTI dataset} } \\
			\textbf{Modalities} \,   & \, $\mathbf{SM_{RGB}}$\,& \, $\mathbf{ML_{RGB}}$ \,& $\mathbf{MAP_{RGB}}$ \,&  $\mathbf{SM_{RV}}$\, & \, $\mathbf{ML_{RV}}$ \,& $\mathbf{MAP_{RV}}$  \\ \hline
			\textbf{F-score}    \,   & \, $95.89$   \,& \, $94.85$    \,& $95.07$     \, &  $90.29$  \, & \, $89.70$   \,& $89.50$ \\
			$\mathbf{FPR}$\,   & \, $1.60$    \,& \, $1.21$     \,& $1.19$      \,&  $2.73$   \, & \, $2.64$    \,& $2.60$  \\
			$\mathbf{Ave.Scores_{FP}}$\,   & \, $0.853$    \,& \, $0.487$     \,& $0.359$      \,&  $0.874$   \, & \, $0.656$    \,& $0.387$\\ 
			$\mathbf{Var.Scores_{FP}}$\,   & \, $0.021$    \,& \, $0.018$     \,& $0.003$      \,&  $0.028$   \, & \, $0.024$    \,& $0.004$  \\\hline
			$\mathbf{Ave.Scores_{unseen}}$  & \, $0.983$ \, & \, $0.708$ \, & \, $0.397$ \, & \, $0.970$ \, & \, $0.692$ \, & \, $0.394$ \\
			$\mathbf{Var.Scores_{unseen}}$  & \, $0.005$  \, & \, $0.025$  \, & \, $0.004$  \, & \, $0.010$  \, & \, $0.017$  \, & \, $0.003$ \\
			\toprule
			\multicolumn{7}{c}{ \textbf{LL5 dataset} } \\
			\textbf{Modalities} \,   & \, $\mathbf{SM_{RGB}}$\,& \, $\mathbf{ML_{RGB}}$ \,& $\mathbf{MAP_{RGB}}$ \,&  $\mathbf{SM_{RV}}$\, & \, $\mathbf{ML_{RV}}$ \,& $\mathbf{MAP_{RV}}$  \\ \hline
			\textbf{F-score}   \,   & \, $92.85$   \,& \, $92.84$    \,& $92.91$     \, &  $90.16$  \, & \, $89.91$   \,& $89.94$ \\
			$\mathbf{FPR}$\,   & \, $2.40$    \,& \, $2.16$     \,& $1.98$      \,&  $2.17$   \, & \, $1.78$    \,& $1.76$  \\ 
			$\mathbf{Ave.Scores_{FP}}$\,   & \, $0.939$    \,& \, $0.531$     \,& $0.383$      \,&  $0.914$   \, & \, $0.574$    \,& $0.398$  \\ 
			$\mathbf{Var.Scores_{FP}}$\,   & \, $0.015$    \,& \, $0.040$     \,& $0.009$      \,&  $0.017$   \, & \, $0.036$    \,& $0.009$  \\\hline
			$\mathbf{Ave.Scores_{unseen}}$  & \, $0.996$ \, & \, $0.454$ \, & \, $0.375$ \, & \, $0.996$ \, & \, $0.502$ \, & \, $0.374$ \\
			$\mathbf{Var.Scores_{unseen}}$  & \, $0.001$  \, & \, $0.037$  \, & \, $0.009$  \, & \, $0.001$  \, & \, $0.038$  \, & \, $0.006$ \\
			\bottomrule
		\end{tabular} \noindent
		\label{result}
	\end{center}
\end{table*} \noindent
\raggedbottom

All classes for the training dataset were extracted directly from the aforementioned datasets, except for the tree, lamppost, and signpost classes which were manually extracted from the data for this study. The rationale behind this is to evaluate the prediction confidence of the network on objects that do not belong to any of the trained classes, and as such the consistency of the models can be assessed. Ideally, if the classifiers are perfectly consistent in terms of probability interpretation, the prediction scores would be identical (equal to $1/3$) for each class in each sample of the unseen dataset. Results on the testing set are shown in Table \ref{result} in terms of F-score, false positive rate ($FPR$), the average ($Ave.Scores_{FP}$) and variance ($Var.Scores_{FP}$) of the false positives ($FP$). The average ($Ave.Scores_{unseen}$) and the variance ($Var.Scores_{unseen}$) of the predicted scores are also shown for the unseen testing set (out-of-training distribution test data).

In reference to Table \ref{result}, where the results are reported based on the classification test set, it can be observed that the $FPR$, $Ave.Scores_{FP}$ and $Var.Scores_{FP}$ values are considerably lower than the results presented by the \textit{SM} layer for both of the sensor modalities and datasets. Regarding the F-scores of the proposed approach (\textit{ML} and \textit{MAP}) compared to the \textit{SM} resulted in an average reduction of $1\%$ (percentage point) for the RGB modality and $0.76\%$ for RV modality, considering KITTI dataset. The F-scores on the LL5 dataset got a gain of $0.065\%$ for RGB modality, considering the \textit{MAP} approach, F-score of the VR modality had a average reduction of $0.26\%$. Such reductions of the F-scores are relatively small and thus did not compromise the classification ability. Additionally, the distribution of the top-label scores on the test set comprising the objects that belong to the trained classes (in-distribution classes) is discussed in the Appendix \ref{AnnexPS}.

\begin{figure*}[!t]
	\centering
	\begin{subfigure}[b]{0.99\textwidth}
		\centering
		\includegraphics[width=\textwidth]{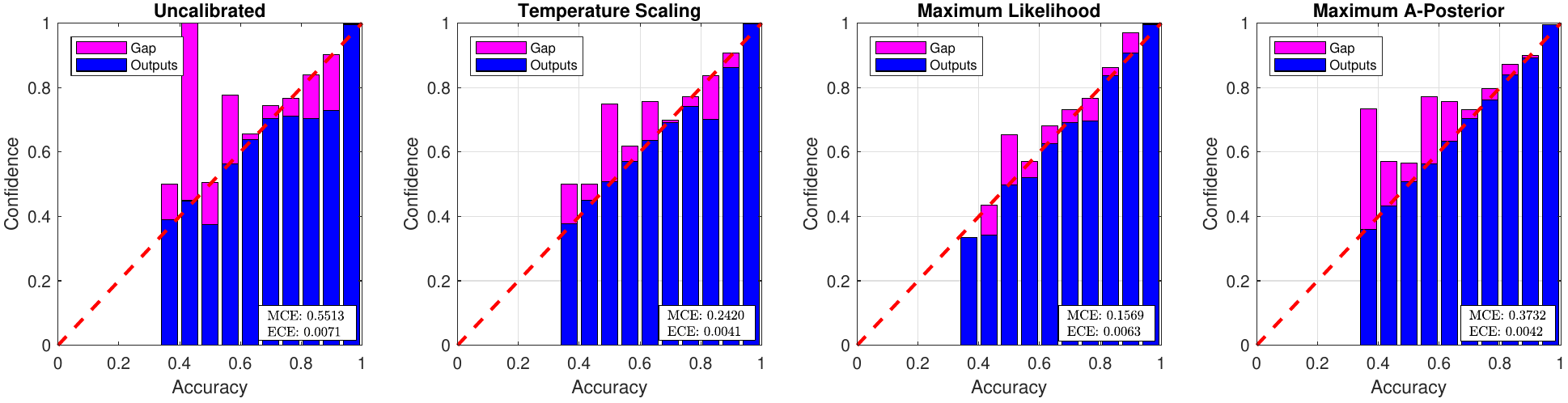}
		\caption{Reliability diagrams for RGB images from KITTI dataset, considering the number of bins $=15$ and $TS=1.31$.}
		\label{RGB_KITTI_RelDiag}
	\end{subfigure}
	\hfill\vspace{0.4cm}
	\begin{subfigure}[b]{0.99\textwidth}
		\centering
		\includegraphics[width=\textwidth]{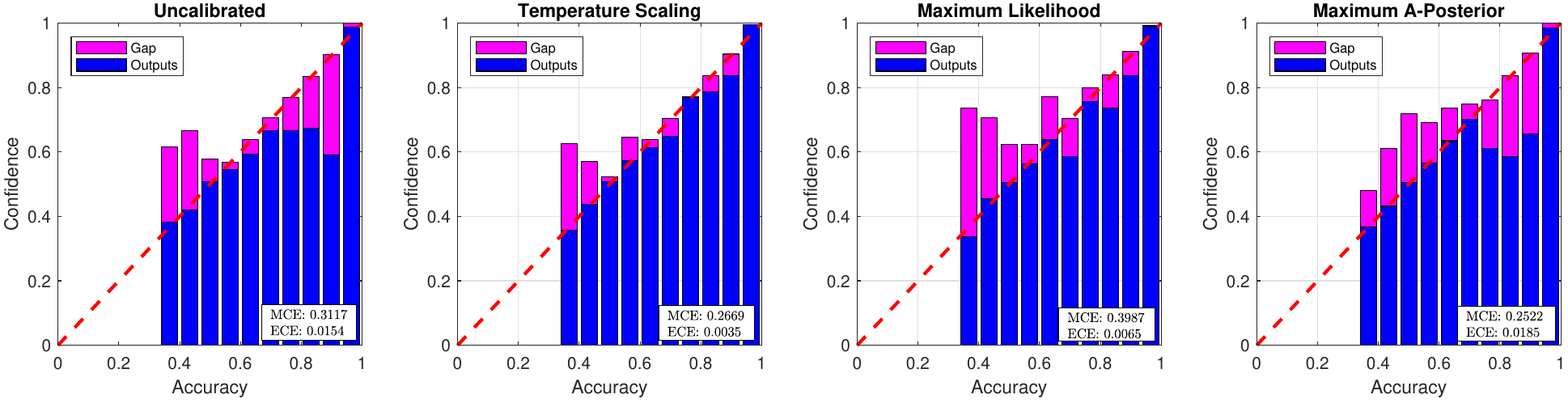}
		\caption{Reliability diagrams for RV images from KITTI dataset, considering the number of bins $=15$ and $TS=2.26$.}
		\label{DM_KITTI_RelDiag}
	\end{subfigure}
	\caption{The graphs, from left to right, represent uncalibrated score values, followed by score values calibrated through Temperature Scaling, then scores obtained by the \textit{ML} and \textit{MAP} layers respectively.}
	\label{KITTI_RelDiag}
\end{figure*}

\begin{figure*}[!t]
	\centering
	\begin{subfigure}[b]{0.99\textwidth}
		\centering
		\includegraphics[width=\textwidth]{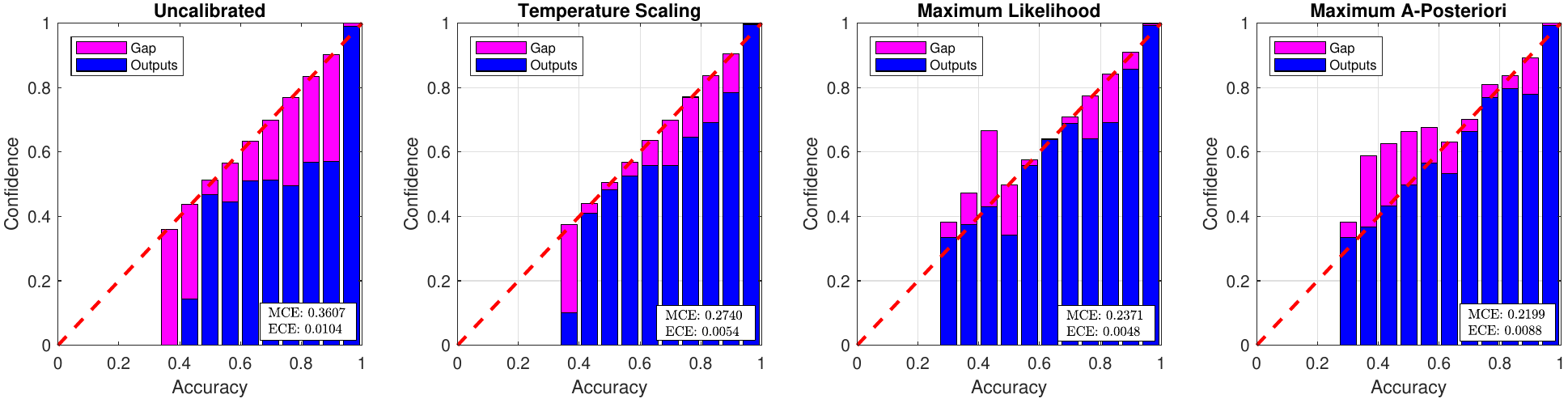}
		\caption{Reliability diagrams for RGB images from Lyft Level 5 dataset, considering the number of bins $=15$ and $TS=2.46$.}
		\label{RGB_LL5_RelDiag}
	\end{subfigure}
	\hfill\vspace{0.4cm}
	\begin{subfigure}[b]{0.99\textwidth}
		\centering
		\includegraphics[width=\textwidth]{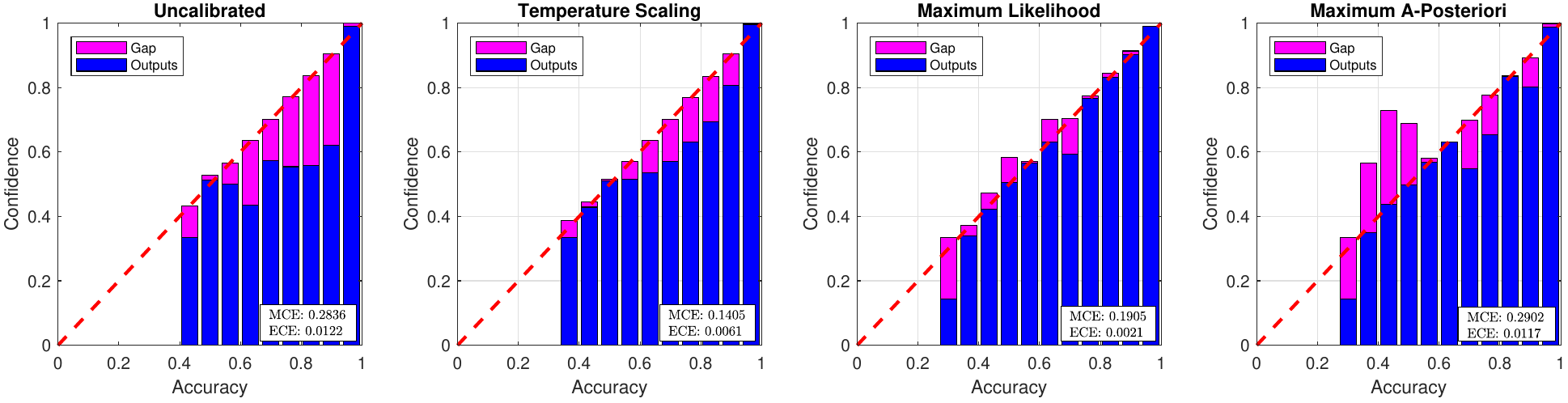}
		\caption{Reliability diagrams for RV images from Lyft Level 5 dataset, considering the number of bins $=15$ and $TS=1.90$.}
		\label{DM_LL5_RelDiag}
	\end{subfigure}
	\caption{Reliability diagrams, on the LL5 dataset, for the following cases (from left-right): uncalibrated scores, calibrated model using TS, and then the diagrams for the models using \textit{ML} and \textit{MAP} layers.}
	\label{LL5_RelDiag}
\end{figure*}

Another way of analyzing the results of reducing overconfidence predictions is through reliability diagrams, as shown in the figures \ref{KITTI_RelDiag} and \ref{LL5_RelDiag}, considering uncalibrated, \textit{ML} and \textit{MAP} data. Furthermore, as a way of validating our methodology, we compared our results achieved with the temperature scaling calibration technique. Note that the results presented through the reliability diagrams are shown through the MCE and ECE metrics. From these metrics we cannot say which is the best calibration technique, because for a given technique the lowest value for the MCE was obtained, while for another technique the lowest value for the ECE was obtained. However, we show that the proposed approach contributed to reduce the calibration errors \ie, to reduce the values of the MCE and ECE metrics when compared to the uncalibrated data, and consequently we provide a more reliable result, as well as the contribution to reduce the overconfidence predictions.
 \noindent

 \noindent

\begin{figure}[!t]
	\begin{subfigure}[t]{0.475\textwidth}
		\centering
		\includegraphics[width=\textwidth]{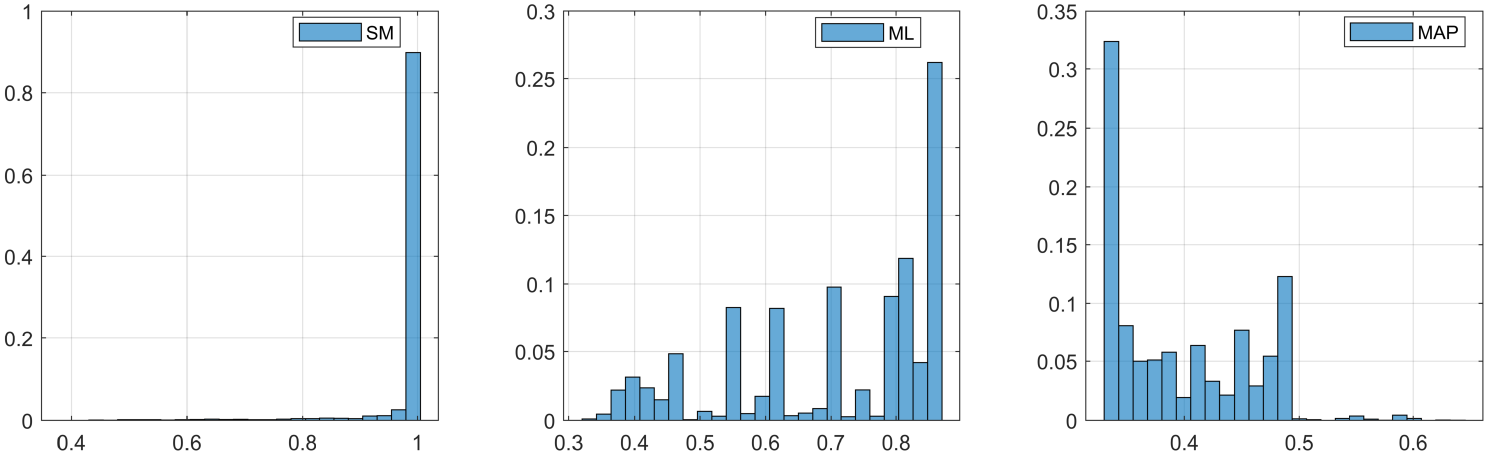}
		\caption{\textit{SM}, \textit{ML}, \textit{MAP} scores on the RGB `unseen' KITTI-set.}
		\label{HG_Softmax_RGB_US}
	\end{subfigure}
	\\
	\begin{subfigure}[t]{0.475\textwidth}
		\centering
		\includegraphics[width=\textwidth]{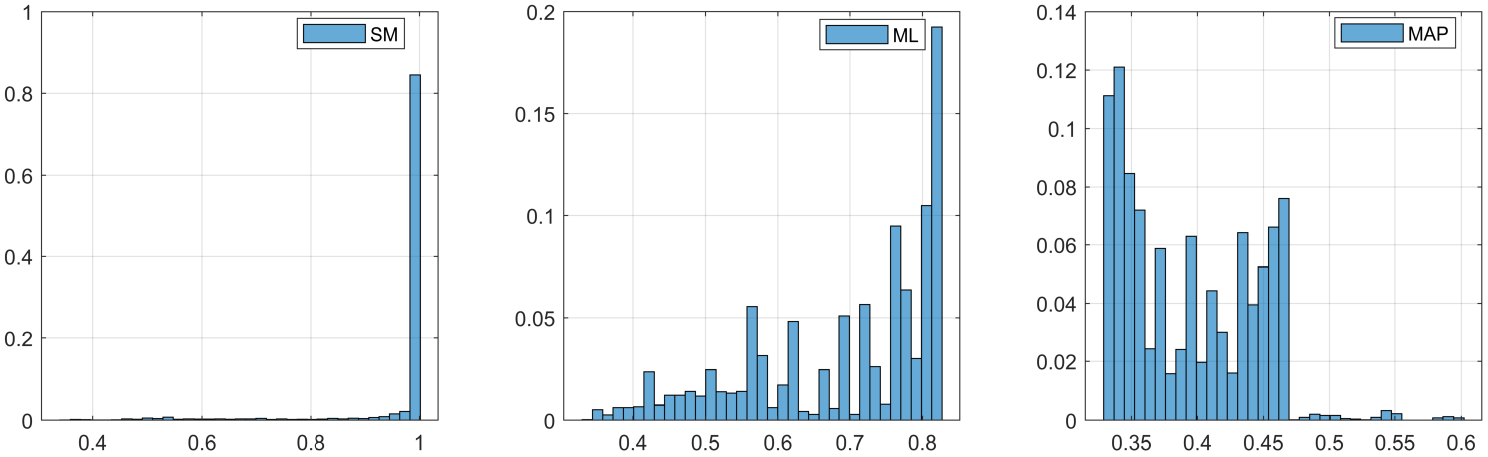}
		\caption{\textit{SM}, \textit{ML}, \textit{MAP} scores on the RV `unseen' KITTI-set.}
		\label{HG_Softmax_DM_US}
	\end{subfigure}
	\begin{subfigure}[t]{0.475\textwidth}
		\centering
		\includegraphics[width=\textwidth]{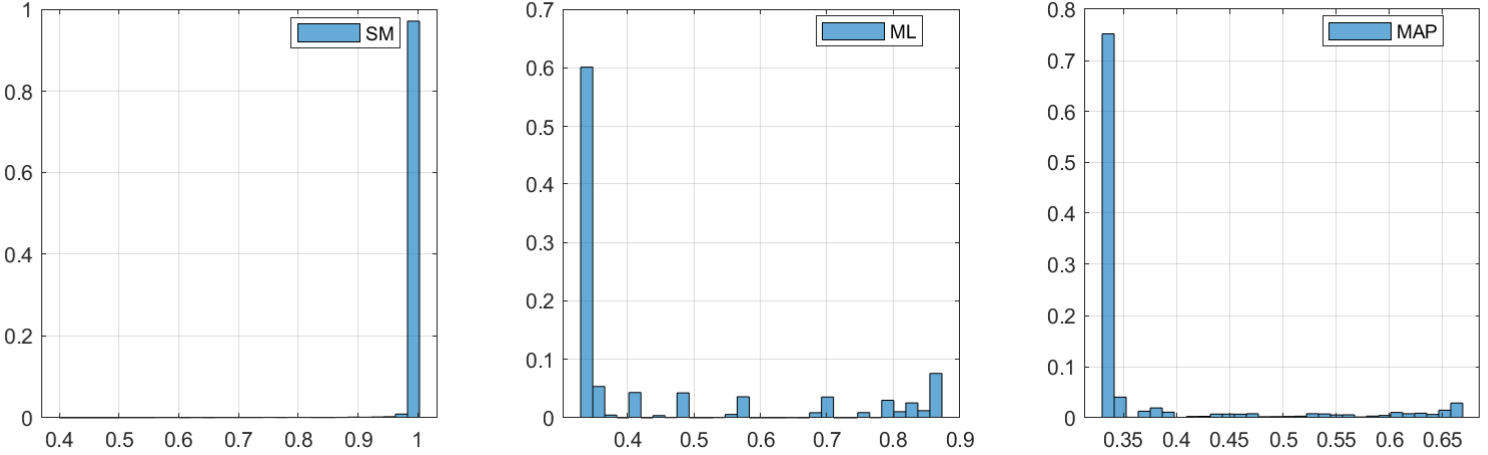}
		\caption{\textit{SM}, \textit{ML}, \textit{MAP} scores on the RGB `unseen' LL5-set.}
		\label{HG_Softmax_RGB_US_LL5}
	\end{subfigure}
	\\
	\begin{subfigure}[t]{0.475\textwidth}
		\centering
		\includegraphics[width=\textwidth]{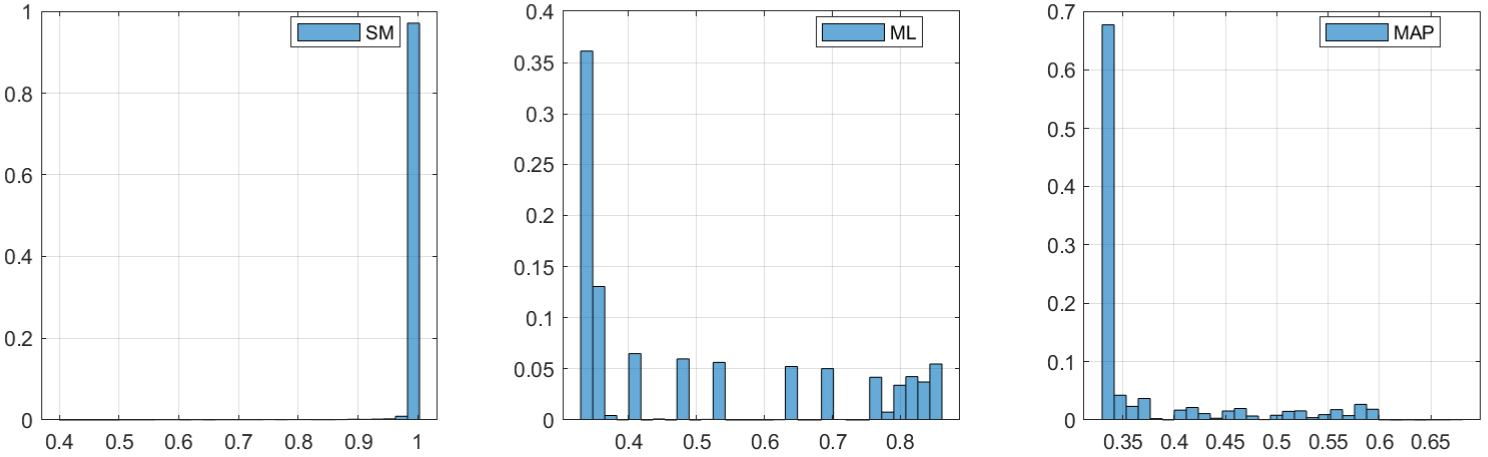}
		\caption{\textit{SM}, \textit{ML}, \textit{MAP} scores on the RV `unseen' LL5-set.}
		\label{HG_Softmax_DM_US_LL5}
	\end{subfigure}
	\caption{Prediction scores on the unseen/non-trained data (comprising the classes: person sitting, tram, tree/ lamppost/signpost, truck, van), using \textit{SM} layer (left side), and the proposed \textit{ML} (center) and \textit{MAP} (right side) layers. The graphs of the first two rows are the results of the KITTI dataset, while the last two are from the LL5 dataset.}
	\label{fig:Unseen}
\end{figure}

Further experiments have been carried out as a complementary analysis concerning the network's overconfidence behaviour, on a so-called `unseen' test set, by means of the network's average score $Ave.Scores_{unseen}$. Note that for \textit{ML} and \textit{MAP}  layers, the results are smaller than the \textit{SM} layer as can be seen in Table \ref{result}. This indicates that the probabilistic inferences are significantly better balanced \ie, enabling more reliable decision-making, when `new' objects of `non-trained' classes are presented to the CNNs, as illustrated by Fig. \ref{fig:Unseen} \ie, the distribution for the unseen dataset. We can see that the aforementioned graphs show less extreme results than those provided by the \textit{SM} layer.

 \noindent

\begin{figure}[!t]
	\begin{subfigure}[t]{0.475\textwidth}
		\centering
		\includegraphics[width=\textwidth]{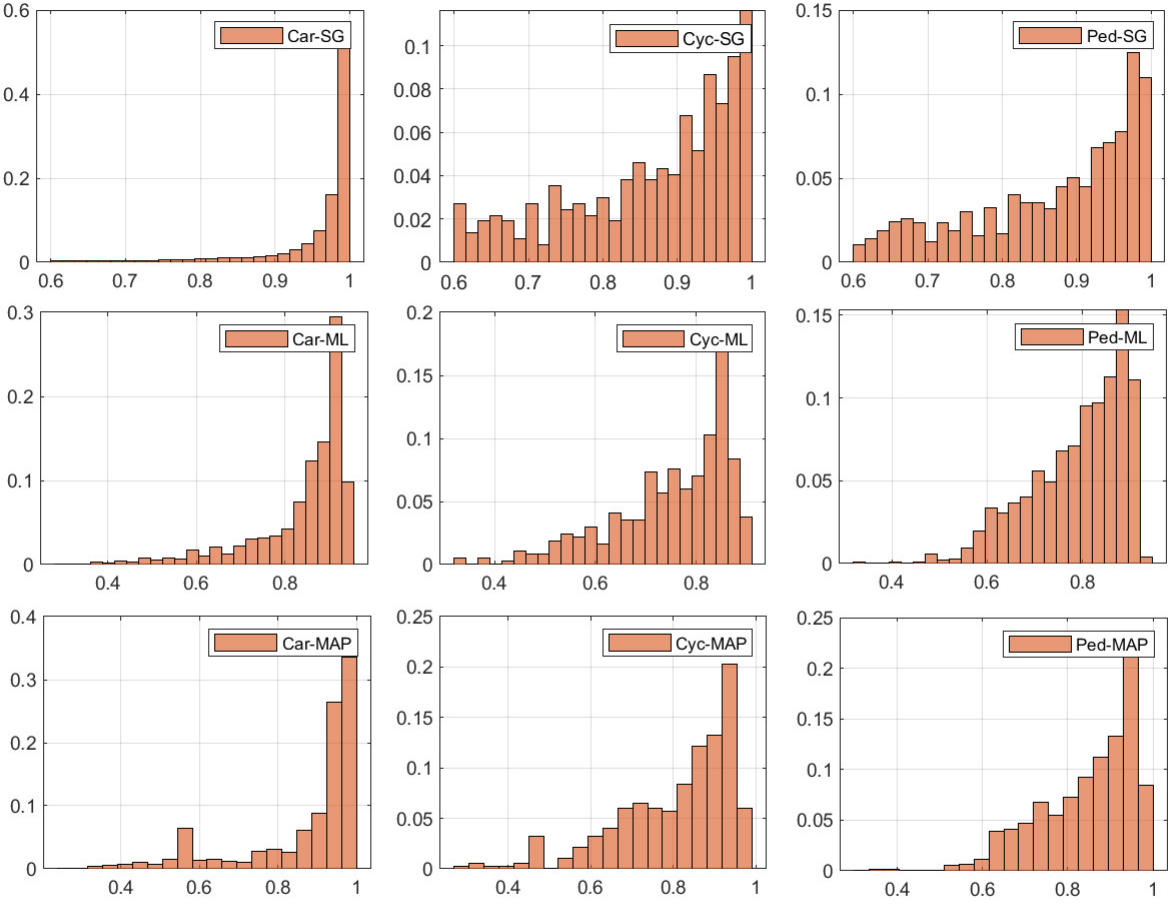}
		\caption{Scores from the true positives.}
		\label{sg_tp}
	\end{subfigure}
	\begin{subfigure}[t]{0.475\textwidth}
		\centering
		\includegraphics[width=\textwidth]{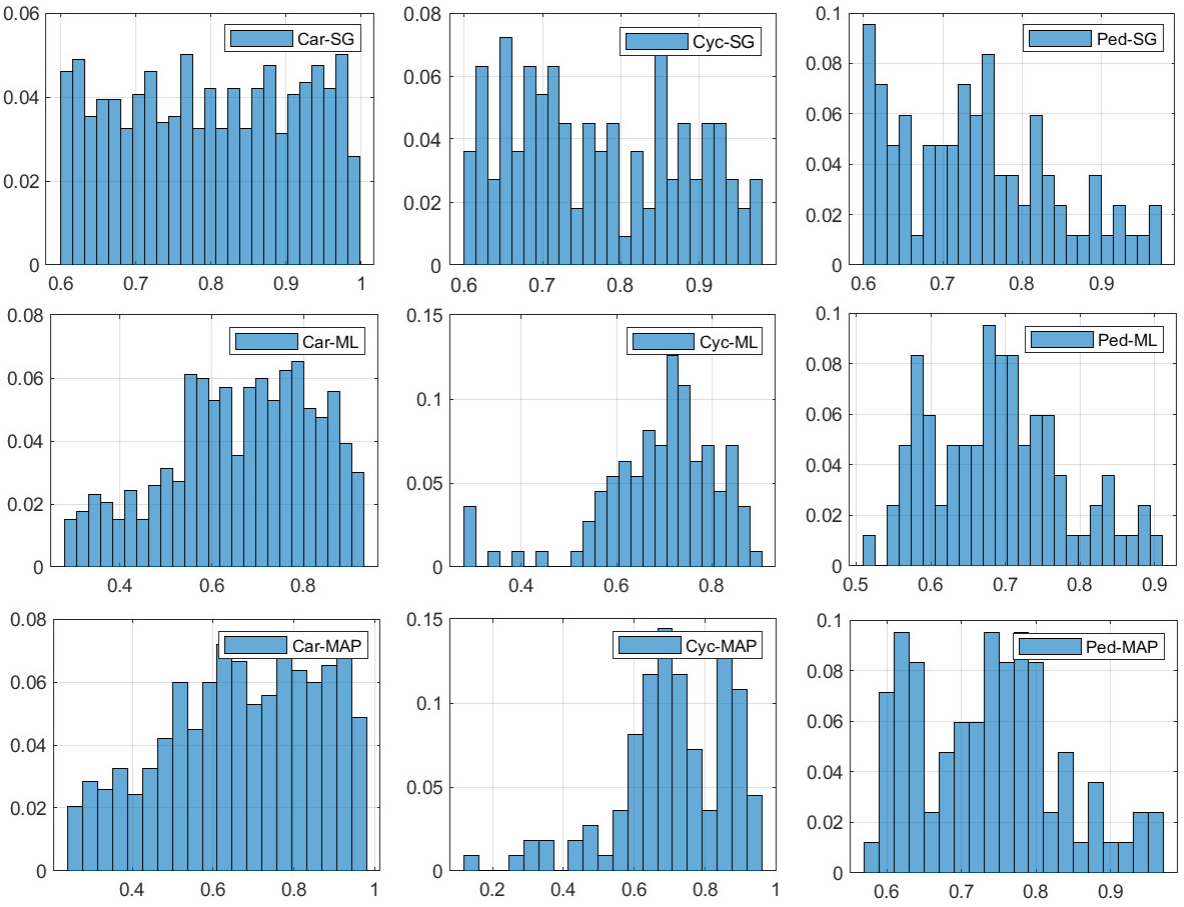}
		\caption{Scores from the false positives.}
		\label{sg_fp}
	\end{subfigure}
	\caption{Results obtained from the Yolo V4. The columns from left to right represent the car, cyclist and pedestrian classes, as well as the distributions of the Sigmoid layer, Maximum Likelihood and Maximum a-Posteriori functions scores. The first line of the distributions are the results of the classifications of the true positives, while the last line is the corresponding scores of the false positives.}
	\label{sg_ml_map_tp}
\end{figure} \noindent

\subsection{Results on Object Detection}

\begin{figure*}[!t]
	\centering
	\includegraphics[scale=0.55]{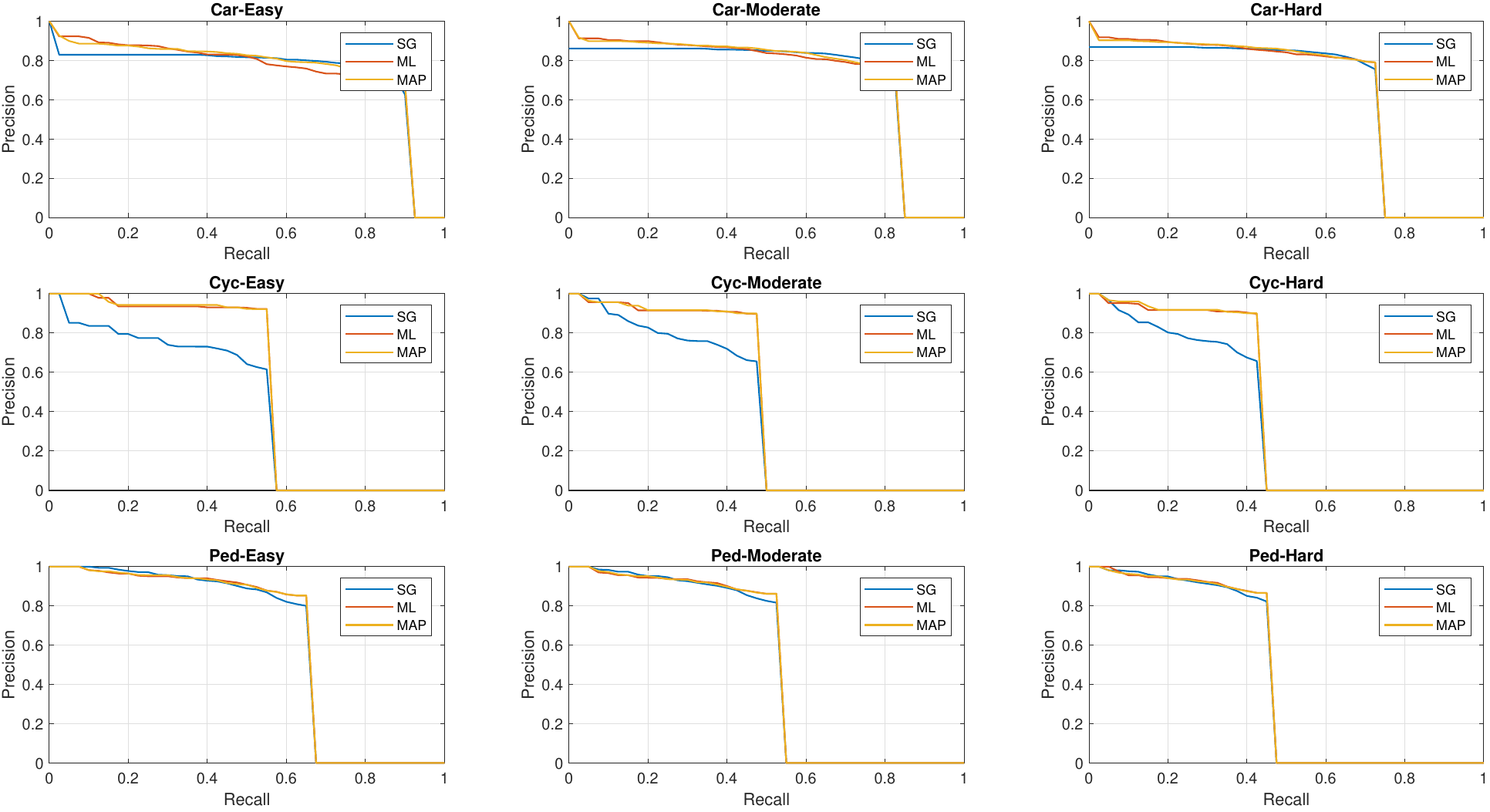}
	\caption{Precision-recall curves for Yolo V4 obtained from the Sigmoid prediction layer, \textit{ML} and \textit{MAP} layers on the KITTI dataset, considering the true positives. The curves were obtained for the easy, moderate and hard cases, according to the toolbox provided by KITTI.}
	\label{precision_recall}
\end{figure*}

The results on the object detection dataset using the \textit{ML} and \textit{MAP}  nonlinearities are impressive. Such results were not presented through reliability diagrams, but through normalized histograms, which showed more clearly the reduction in overconfidence in relation to objects detected as false positives without degrading the results of the true positives, as showed in Fig. \ref{sg_ml_map_tp}. The results are more representative through precision-recall curves, especially for the cyclist class (Cyc), whose areas under the curves (AUCs) are $24.03\%$, $14.28\% $ and $14.63\% $ for the easy, moderate and hard cases respectively, as shown in Fig. \ref{precision_recall} and Table \ref{auc}. With respect to the car (Car) and pedestrians (Ped), the proposed approach also showed some improvement.
\begin{table*}[!t]
	\begin{center}
		\caption{Comparison of the areas under the curves ($\%$) between the Sigmoid layer (SG), \textit{ML} and \textit{MAP}  layers from the precision-recall curves.}
		\begin{tabular}{cccc|cccc|cccc}
			\hline \hline
			\multicolumn{4}{c}{\textbf{Easy}}          & \multicolumn{4}{c}{\textbf{Moderate}}        & \multicolumn{4}{c}{\textbf{Hard}}\\ 
			\hline \hline
			& \textbf{SG}      & \textbf{ML}      & \textbf{MAP}     &  & \textbf{SG}      & \textbf{ML}      & \textbf{MAP}     &  & \textbf{SG}      & \textbf{ML}      & \textbf{MAP}     \\
			\textbf{Car}   & $73.62$ & $74.29$ & $75.18$ & \textbf{Car}   & $70.47$ & $71.34$ & $71.68$ & \textbf{Car}   & $62.74$ & $63.77$ & $63.85$ \\
			\textbf{Cyc}   & $43.24$ & $53.43$ & $53.63$ & \textbf{Cyc}   & $39.70$ & $45.31$ & $45.37$ & \textbf{Cyc}   & $35.61$ & $40.62$ & $40.82$ \\
			\textbf{Ped}   & $61.82$ & $62.07$ & $62.08$ & \textbf{Ped}   & $49.79$ & $50.01$ & $50.03$ & \textbf{Ped}   & $42.90$ & $43.14$ & $43.08$ \\
			\hline \hline
		\end{tabular}
		\label{auc}
	\end{center}
\end{table*} \noindent
\raggedbottom

 \noindent

Note that the proposed methodology is dependent on the number of bins ($nbins$) and the parameter $\lambda$. Thus, the values of the scores may vary according to the values of these parameters. For the particular case of the cyclist class, the proposed methodology achieved strong classification performance compared to the baseline (results in Table \ref{auc}). In this paper we have chosen to use a single set of parameters for all the three cases (\ie, the same values of $\lambda$ and $nbins$ for each class). Given the proposed approach, we note that a set of tailored parameters for each class can be used instead, as the distributions (PDF's) are carried out individually.

\section{Discussion and Conclusion}
\label{sec:conclusions}

Within the experiments performed in this work, a probabilistic approach for CNNs was addressed as distributions in the Logit layer to better represent the classification outputs. The results reported within the experiments in this work are promising given that \textit{ML} and \textit{MAP}  noticeably reduced the classifier overconfidence and provided a more significant distribution in terms of probabilistic interpretation. 

The improvement is not as significant when analyzing objects defined as true positives. But, our concern is to develop a methodology that can reduce the values of false positives (mainly objects of the unseen class: which may be critical in robotics and autonomous driving applications) without degrading the results achieved by true positives. Note that we have included two metrics in Table \ref{result}, in order to show the reduction of score values for the `unseen' class (in particular) and also to show that the overconfidence behavior has been mitigated for TPs and FPs.

One potential way to improve the F-scores achieved by the \textit{ML} and \textit{MAP} layers would be to obtain a `perfect' match between the smoothing parameter ($\lambda$) and the number of bins in the histograms. For the new results with the EfficientNetB1 network, we have selected the parameters by using an exhaustive search process (combining several values as possible), in order to keep the values of the F-scores of the \textit{ML} and \textit{MAP} layers practically identical to those achieved by the EfficientNetB1 baseline. Figures \ref{eff_1}, \ref{eff_2}, and \ref{eff_3} show reductions on the scores for objects of class `unseen' thus, the proposed approach is efficient. 

As a consequence of the Additive Smoothing, the score values equal to $0.0$ and $1.0$ are excluded from the prediction values. The influence of the $\lambda$ parameter on the data distribution can be seen from the figures in Appendix \ref{AnnexSP}, particularly with respect to objects of the `unseen' class.

To assess the classifier's robustness or the uncertainty of the model when predicting objects of unseen classes by the network, we considered a test set comprised of `new' objects. Overall, the results are promising, since the distribution of the predictions were not extremities relative to the results from the \textit{SM} layer, in other words, the average scores using \textit{ML} and \textit{MAP}  layers were significantly lower than the Softmax prediction layer (the baseline), and thus the CNNs are less prone to overconfidence.

The results for object classification were presented through reliability diagrams, taking into account the MCE and ECE metrics. In fact, such metrics indicate how much the predicted score values are calibrated, that is, the best calibration has to present the lowest value for the MCE and ECE. However, we observed that depending on the dataset and sensor modality, our approach obtained the best result in only one of the metrics \ie, either the lowest value for the MCE metric or the lowest value for the ECE metric. This fact can also be noticed with the temperature scaling calibration technique.

Another important factor that contributes to validate the proposed approach is the use of two different datasets, in terms of both RGB and Range-View ($3D$ point clouds-LiDARs) modalities, since the sensors of the datasets have different resolutions, mainly the LiDAR sensor; While the KITTI dataset provides $3D$ point clouds obtained from a sensor with 64 beams, the LL5 dataset provides $3D$ point clouds with 40 beams - and so, the proposed approach was also successful with differing sensor resolutions within the state of the art. 

The proposed methodology also obtained good results for object detection, not degrading the results when compared to the \textit{SG} prediction layer, presenting better results in all cases. The improvement is more evident for the `cyclist' class, which contains the least amount of examples. This is an interesting result that could be further investigated in future work.

Regarding the formulations of probabilistic distributions, the prior modeling by a Gaussian distribution was shown to guarantee a smoother distribution for the prediction values. Unlike the prior, the likelihood function was modeled by means of a normalized histogram \ie, by a non-parametric formulation showing the probability distributions. If both the prior and the likelihood function were modeled by a uniform distribution, the final result would be similar to those achieved by the \textit{SM} and \textit{SG} layers, since it would not offer any smoothing for the prediction values. In fact, a uniform prior or likelihood would add a constant to the training data modeling, which would have little effect on the prediction values obtained by the \textit{ML} and \textit{MAP} .

\section{Future Work}
\label{futurework}
Softmax and Sigmoid layers represent confidence measures, but they do not provide any measure of uncertainty of the predictions. In other words, both layers mentioned previously provide a direct measure of certainty through the maximum class probability. Such layers also do not provide any information about the certainty that the model itself has about the predictions. Therefore, we address the issues of overconfident predictions and calibration techniques in this work with a focus on perception systems for autonomous vehicles. However, we realize that there is a lack of studies on how to quantify the certainty/uncertainty of predictions in relation to calibration techniques and reliability diagrams. As we verified that the MCE and ECE metrics that quantify the calibrated data through the reliability diagrams depend on the number of bins of such diagrams, that is, by changing the number of bins, the MCE and ECE metrics can provide new error values. Thus, what is the correct value of bins to ensure that a set of predictions is well calibrated?

Regardless of the methodology to reduce overconfidence predictions or capture uncertainty in predictions, how should we assess the quality of estimated uncertainty independent of calibration and regularization techniques?

Faced with such questions and based on the studies presented in the literature on computing uncertainties of predictions and of calibration and regularization techniques, we found that evaluating the quality of uncertainty estimates is still a challenge for the following reasons:
\begin{itemize}
	\item uncertainty estimates depend on methods, which are performed by means of approximations \ie, by means of inferences;
	\item uncertainty estimates depend on the sample size \ie, the sample size can provide a certain degree of confidence that such a sample is representative;
	\item it is not easy to obtain a ground truth about uncertainty estimates. In fact, during our study we did not verify the ground truth about uncertainty estimates;
	\item study and evaluate the quality of quantitative uncertainty metrics, such as entropy, Mutual Information, Kullback-Leibler Divergence, and predictive variance.
\end{itemize}

Based on the issues mentioned above, we intend to advance research on the quality of uncertainty estimates, including the formulation of reliability diagrams, as a way to quantify the quality of uncertainty estimates.

\section{Appendix}
\label{Annex}
\subsection{Prediction Scores of the Objects on the Testing Set}
\label{AnnexPS}
\begin{figure}[!t]
	\centering
	\begin{subfigure}[b]{0.48\textwidth}
		\centering
		\includegraphics[width=\textwidth]{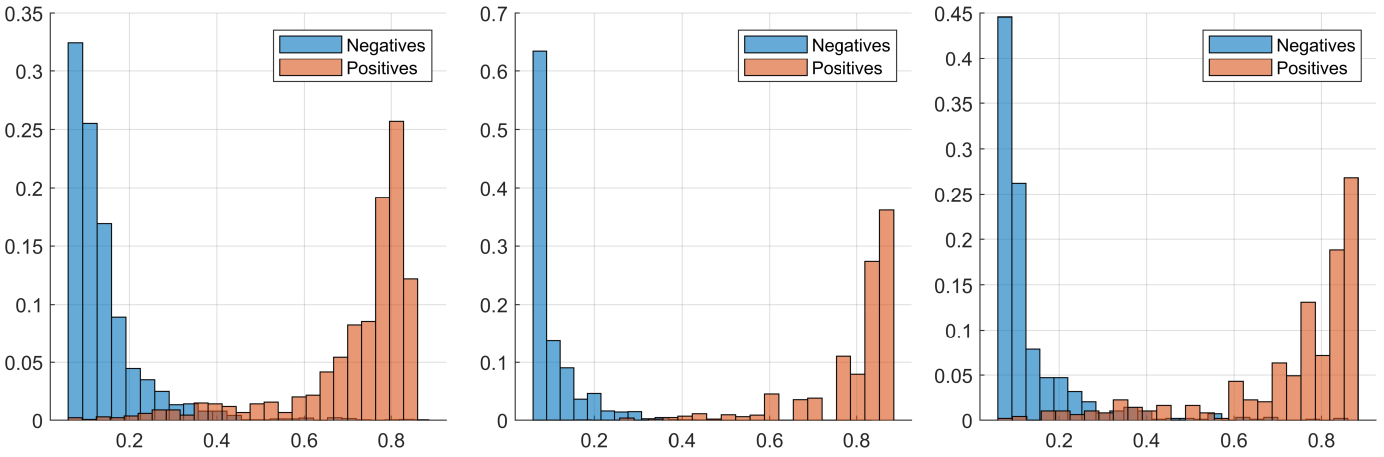}
		\caption{\textit{ML} function scores: RGB.}
		\label{PDF_ML_RGB}
	\end{subfigure}
	\hfill
	\begin{subfigure}[b]{0.48\textwidth}
		\centering
		\includegraphics[width=\textwidth]{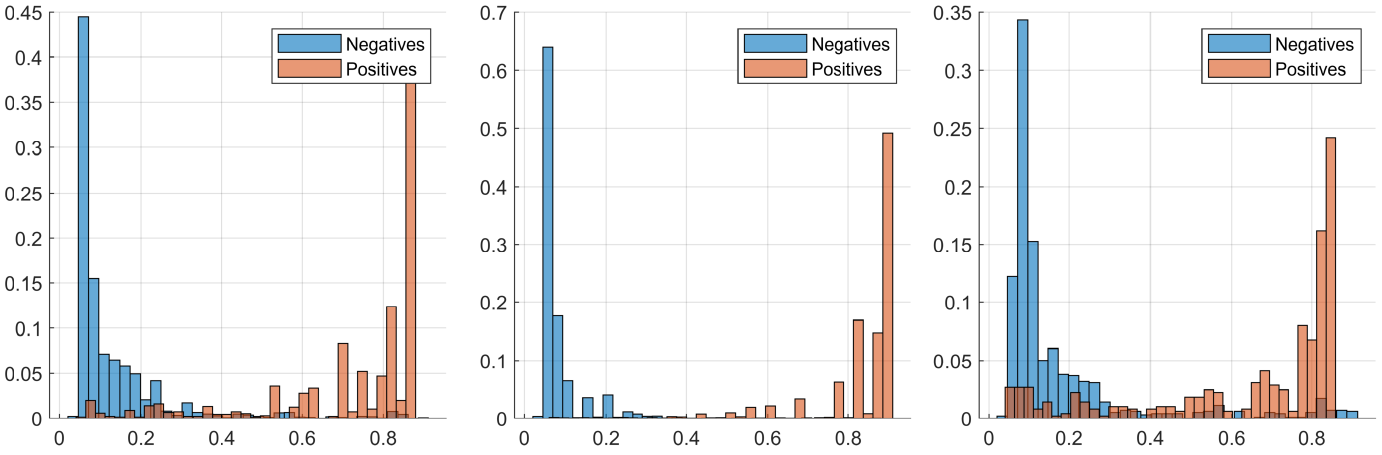}
		\caption{\textit{ML} function scores: RV.}
		\label{PDF_ML_DM}
	\end{subfigure}
	\\
	\begin{subfigure}[b]{0.48\textwidth}
		\centering
		\includegraphics[width=\textwidth]{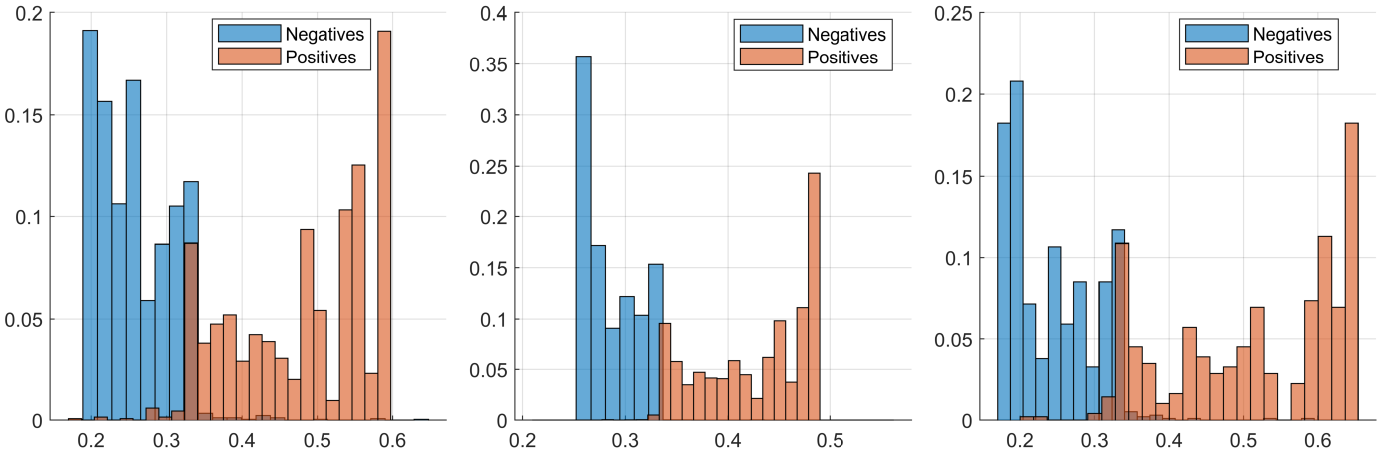}
		\caption{\textit{MAP} function scores: RGB.}
		\label{MAP_RGB}
	\end{subfigure}
	\hfill
	\begin{subfigure}[b]{0.48\textwidth}
		\centering
		\includegraphics[width=\textwidth]{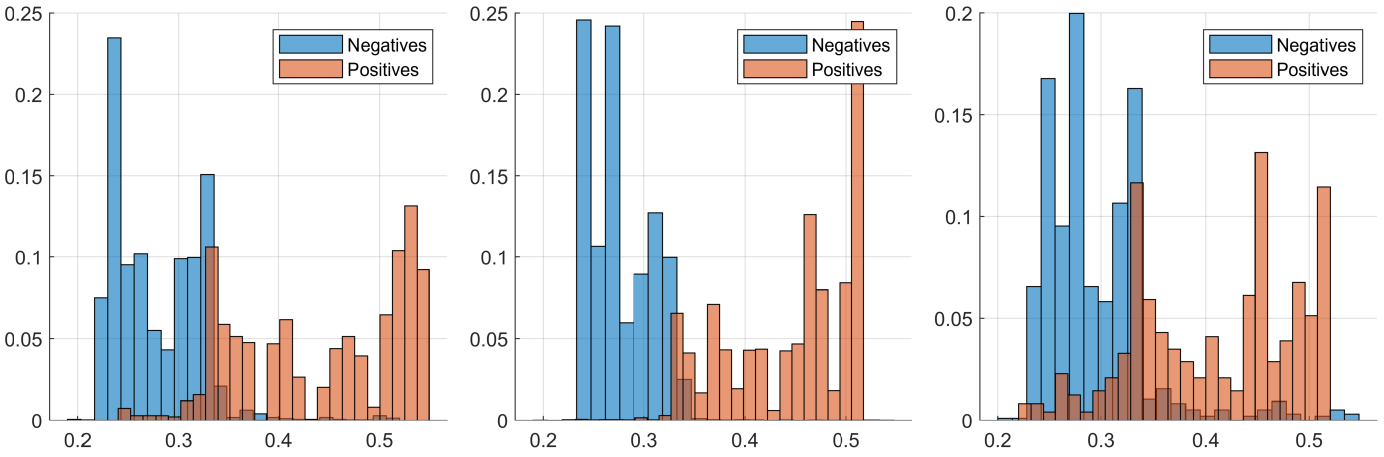}
		\caption{\textit{MAP} function scores: RV.}
		\label{MAP_DM}
	\end{subfigure}
	\caption{From the RGB and LiDAR (RV) modalities, the prediction scores were calculated using the \textit{ML} and \textit{MAP} functions on the KITTI dataset. }
	\label{PDF_MP_MAP_Layer}
\end{figure}

\begin{figure}[!t]
	\centering
	\begin{subfigure}[b]{0.48\textwidth}
		\centering
		\includegraphics[width=\textwidth]{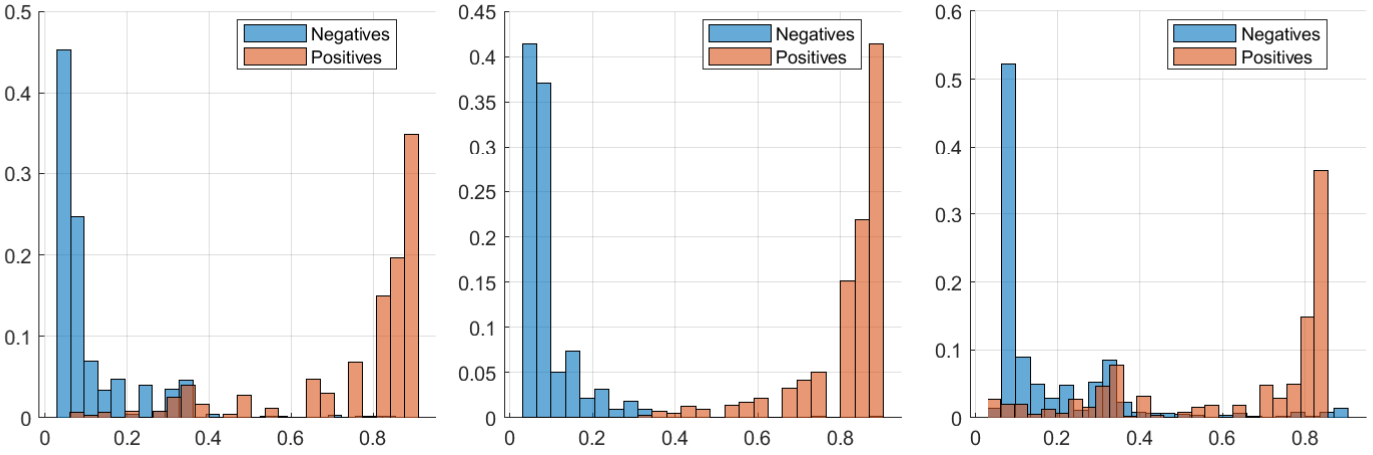}
		\caption{\textit{ML} function scores: RGB.}
		\label{PDF_ML_RGB_LL5}
	\end{subfigure}
	\hfill
	\begin{subfigure}[b]{0.48\textwidth}
		\centering
		\includegraphics[width=\textwidth]{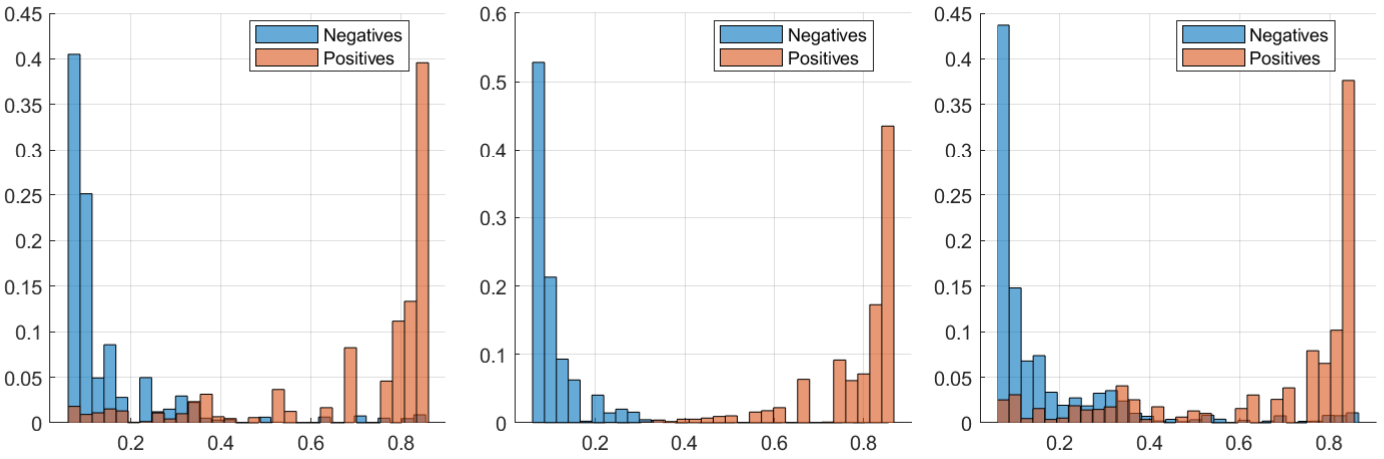}
		\caption{\textit{ML} function scores: RV.}
		\label{PDF_ML_DM_LL5}
	\end{subfigure}
	\\
	\begin{subfigure}[b]{0.48\textwidth}
		\centering
		\includegraphics[width=\textwidth]{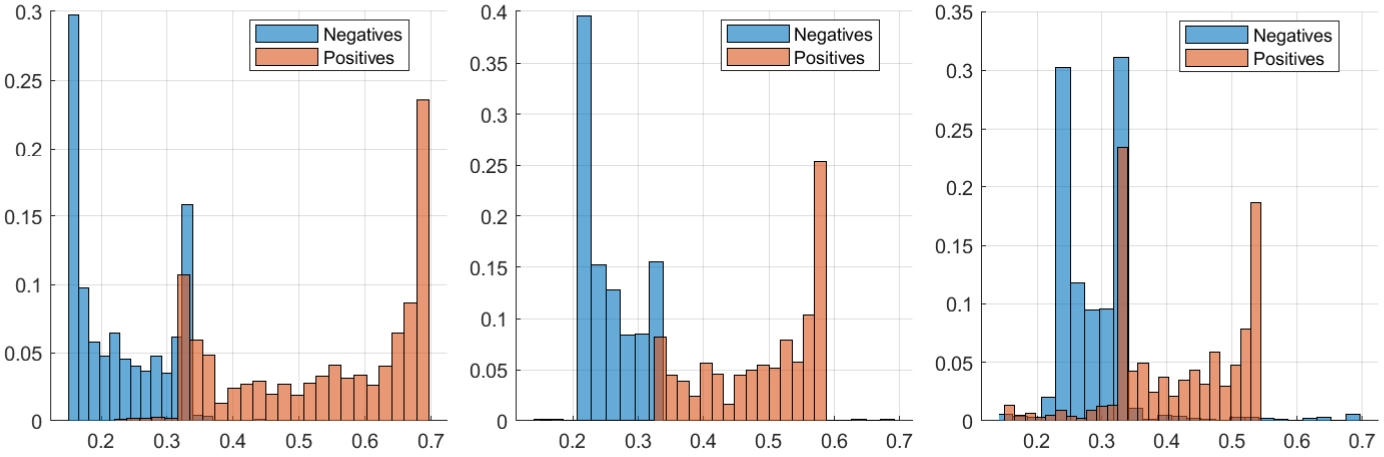}
		\caption{\textit{MAP} function scores: RGB.}
		\label{MAP_RGB_LL5}
	\end{subfigure}
	\hfill
	\begin{subfigure}[b]{0.48\textwidth}
		\centering
		\includegraphics[width=\textwidth]{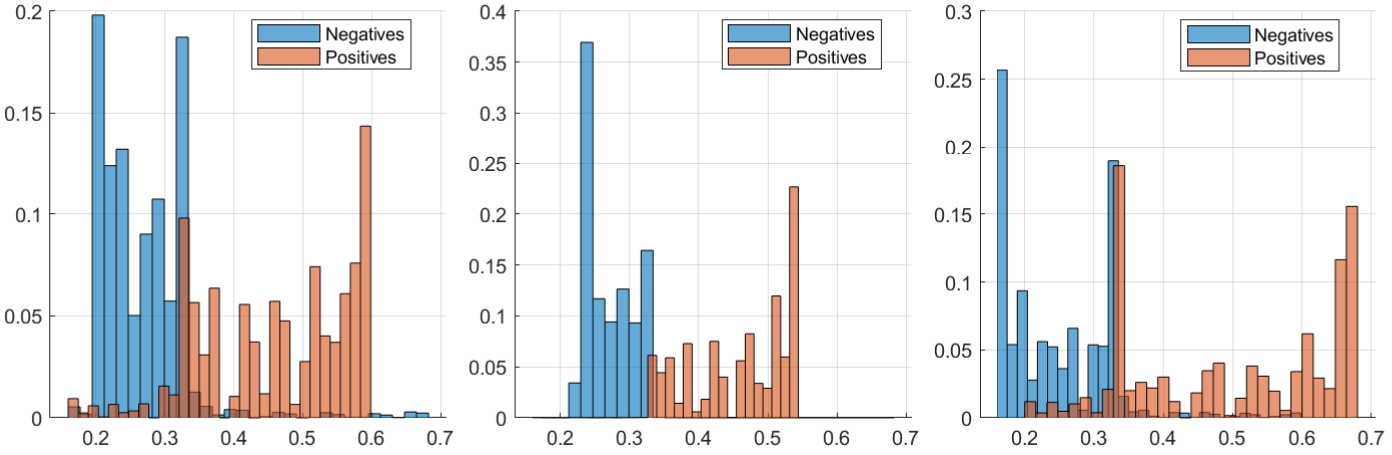}
		\caption{\textit{MAP} function scores: RV.}
		\label{MAP_DM_LL5}
	\end{subfigure}
	\caption{Prediction scores on the testing set for RGB and LiDAR (RV) modalities with the LL5 dataset, using the \textit{ML} and \textit{MAP} functions.}
	\label{PDF_MP_MAP_Layer_LL5}
\end{figure}

The proposed methodology, which is based on the \textit{ML}/\textit{MAP} layers, aims to reduce overconfidence predictions of deep models, especially for objects classified as false positives which sometimes receive high score values of deep networks. An ideal result would be for the network to provide lower score values for the false positives \ie, objects misclassified by the network, and concurrently to attain higher scores for the true positives. As a way of validating additional results on test sets, we present the Fig. \ref{PDF_MP_MAP_Layer} and Fig. \ref{PDF_MP_MAP_Layer_LL5} that contain the results for the pedestrian, car, and cyclist classes (columns from left to right), considering the scores of the objects as being positive and negative, which show smoother distributions of scores when compared to the results shown in Fig. \ref{Softmax}.
 \noindent

 \noindent

\subsection{Smoothing Parameter Influence}
\label{AnnexSP}
\begin{figure}[!t]
	\centering
	\begin{subfigure}[b]{0.45\textwidth}
		\centering
		\includegraphics[scale=0.397]{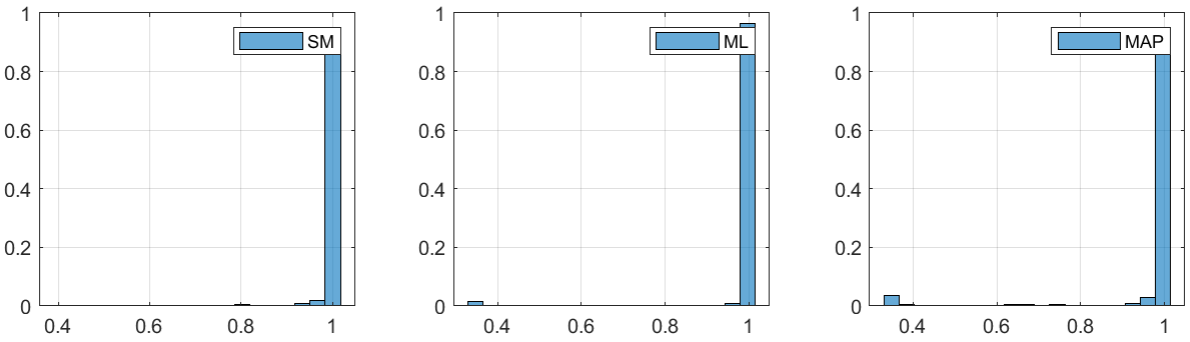}
		\caption{$\lambda_{ML}=1.0\times 10^{-6}$ and $\lambda_{MAP}=1.0\times 10^{-6}$.}
		\label{eff_1_1}
	\end{subfigure}
	\begin{subfigure}[b]{0.45\textwidth}
		\centering
		\includegraphics[scale=0.397]{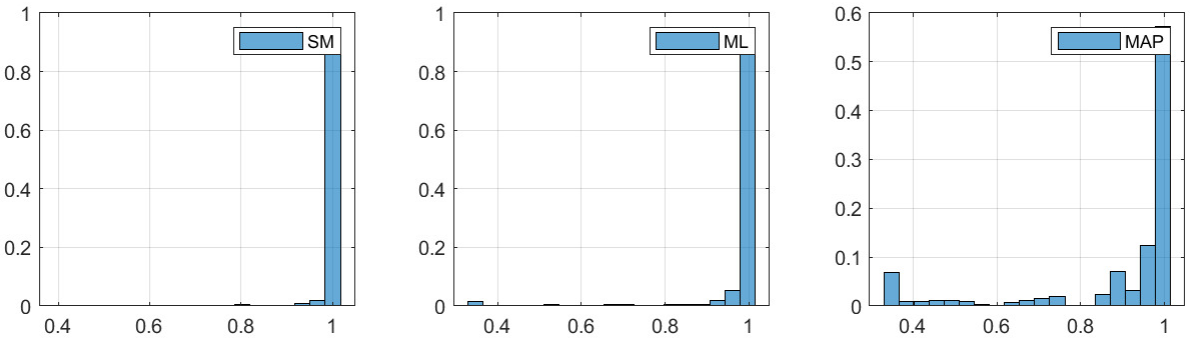}
		\caption{$\lambda_{ML}=9.1\times 10^{-5}$ and $\lambda_{MAP}=9.1\times 10^{-5}$.}
		\label{eff_1_3}
	\end{subfigure}
	\begin{subfigure}[b]{0.45\textwidth}
		\centering
		\includegraphics[scale=0.397]{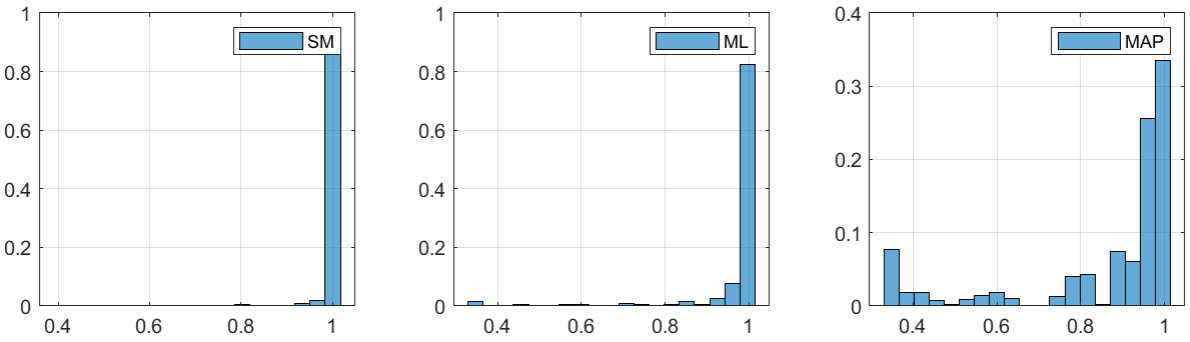}
		\caption{$\lambda_{ML}=1.91\times 10^{-4}$ and $\lambda_{MAP}=1.91\times 10^{-4}$.}
		\label{eff_1_5}
	\end{subfigure}
	\begin{subfigure}[b]{0.45\textwidth}
		\centering
		\includegraphics[scale=0.397]{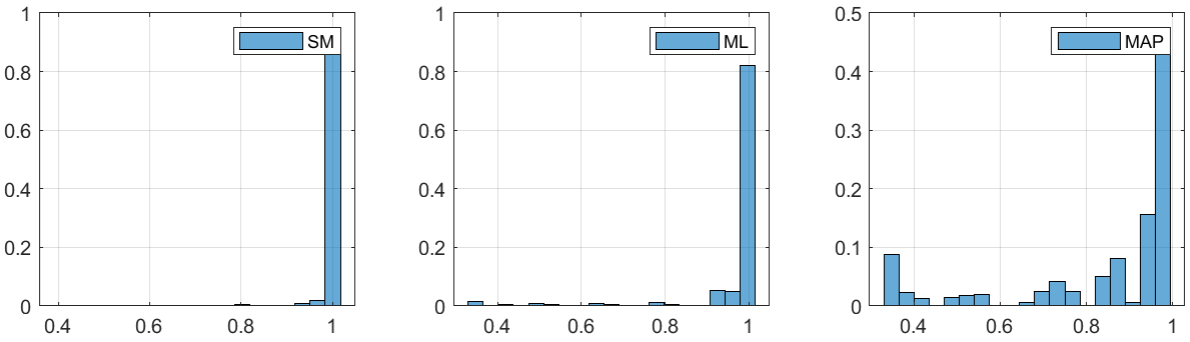}
		\caption{$\lambda_{ML}=2.91\times 10^{-4}$ and $\lambda_{MAP}=2.91\times 10^{-4}$.}
		\label{eff_1_7}
	\end{subfigure}
	\begin{subfigure}[b]{0.45\textwidth}
		\centering
		\includegraphics[scale=0.397]{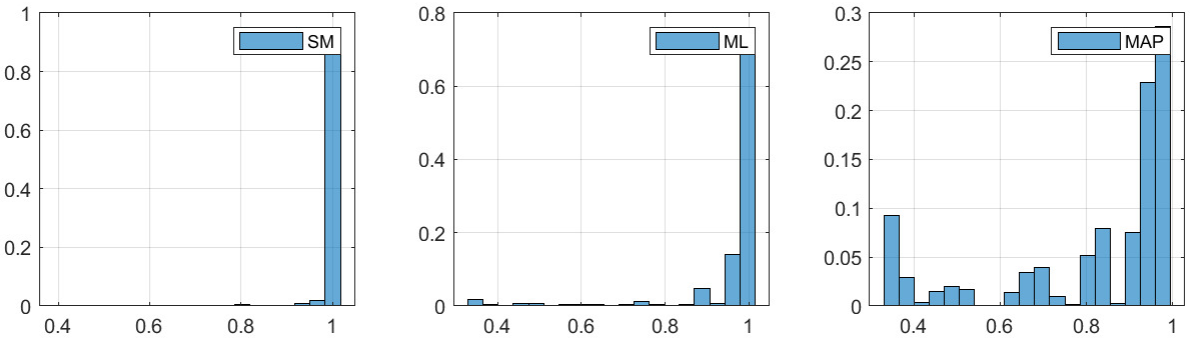}
		\caption{$\lambda_{ML}=3.91\times 10^{-4}$ and $\lambda_{MAP}=3.91\times 10^{-4}$.}
		\label{eff_1_9}
	\end{subfigure}
	\caption{Prediction scores on the RGB unseen/non-trained data, using \textit{SM} layer (left side), and the proposed \textit{ML} (center) and \textit{MAP} (right side). The \textit{SM} case, that does not depend on $\lambda$, serves as baseline for comparison.}
	\label{eff_1}
\end{figure}

\begin{figure}[!t]
	\centering
	\begin{subfigure}[b]{0.45\textwidth}
		\centering
		\includegraphics[scale=0.397]{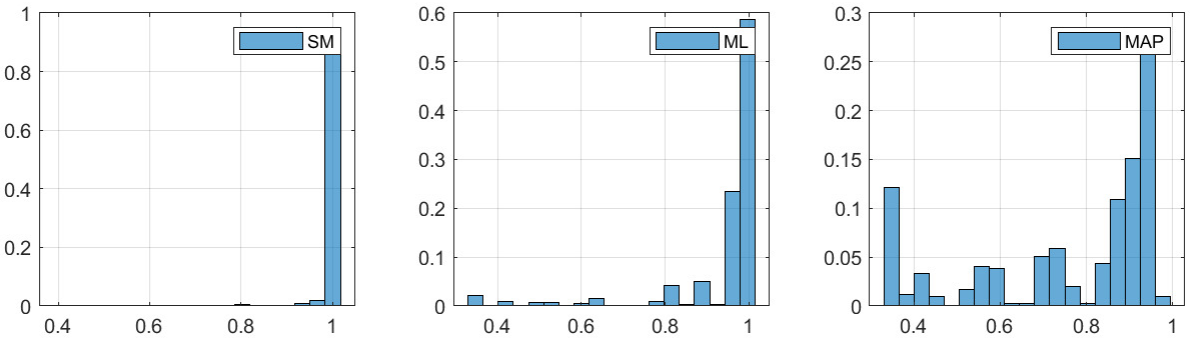}
		\caption{$\lambda_{ML}=7.91\times 10^{-3}$ and $\lambda_{MAP}=7.91\times 10^{-3}$.}
		\label{eff_2_1}
	\end{subfigure}
	\begin{subfigure}[b]{0.45\textwidth}
		\centering
		\includegraphics[scale=0.397]{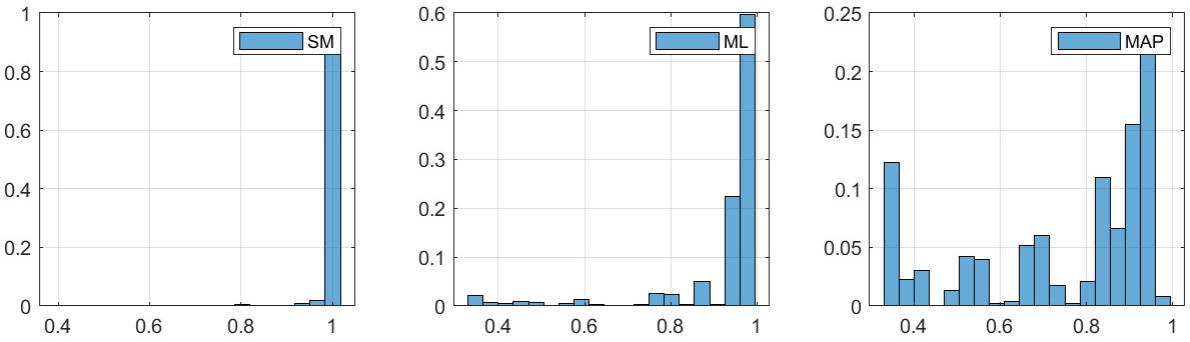}
		\caption{$\lambda_{ML}=9.91\times 10^{-3}$ and $\lambda_{MAP}=9.91\times 10^{-3}$.}
		\label{eff_2_3}
	\end{subfigure}
	\begin{subfigure}[b]{0.45\textwidth}
		\centering
		\includegraphics[scale=0.397]{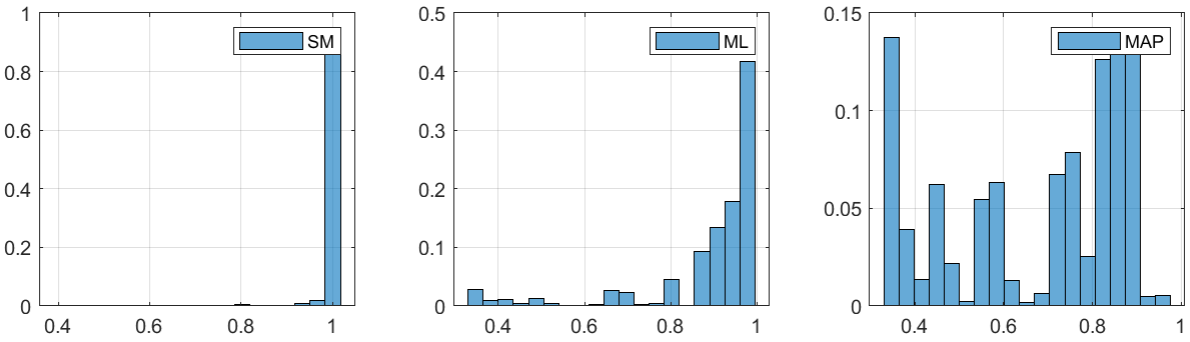}
		\caption{$\lambda_{ML}=1.991\times 10^{-3}$ and $\lambda_{MAP}=1.991\times 10^{-3}$.}
		\label{eff_2_5}
	\end{subfigure}
	\begin{subfigure}[b]{0.45\textwidth}
		\centering
		\includegraphics[scale=0.397]{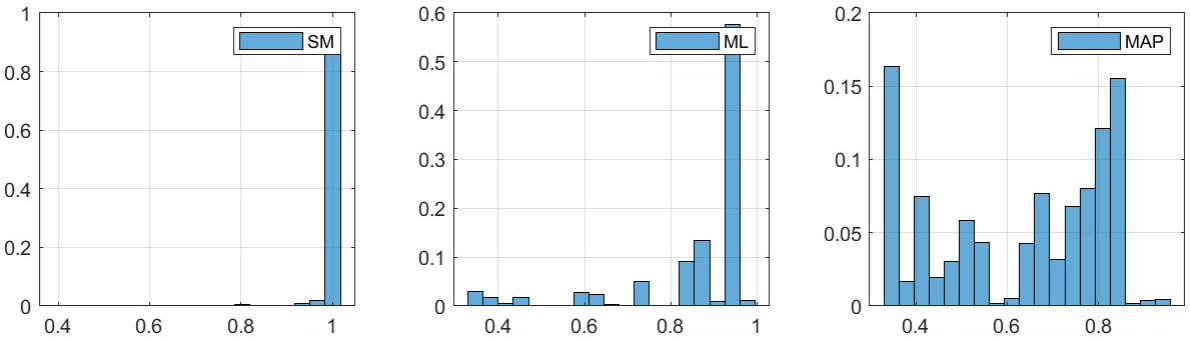}
		\caption{$\lambda_{ML}=2.991\times 10^{-3}$ and $\lambda_{MAP}=2.991\times 10^{-3}$.}
		\label{eff_2_7}
	\end{subfigure}
	\begin{subfigure}[b]{0.45\textwidth}
		\centering
		\includegraphics[scale=0.397]{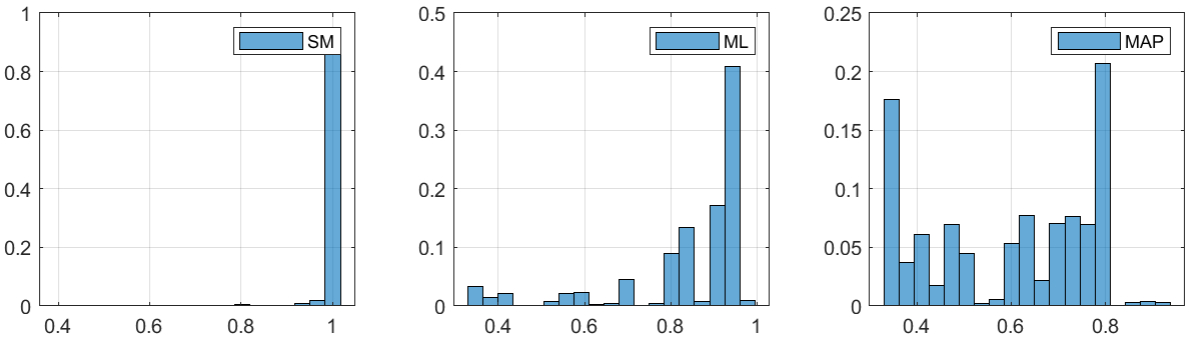}
		\caption{$\lambda_{ML}=3.991\times 10^{-3}$ and $\lambda_{MAP}=3.991\times 10^{-3}$.}
		\label{eff_2_9}
	\end{subfigure}
	\caption{Prediction scores on the unseen data (RGB modality), for the \textit{SM} layer (left side), and the variations in the \textit{ML} (center) and \textit{MAP} layers for different values of $\lambda$.}
	\label{eff_2}
\end{figure}

\begin{figure}[!t]
	\centering
	\begin{subfigure}[b]{0.45\textwidth}
		\centering
		\includegraphics[scale=0.397]{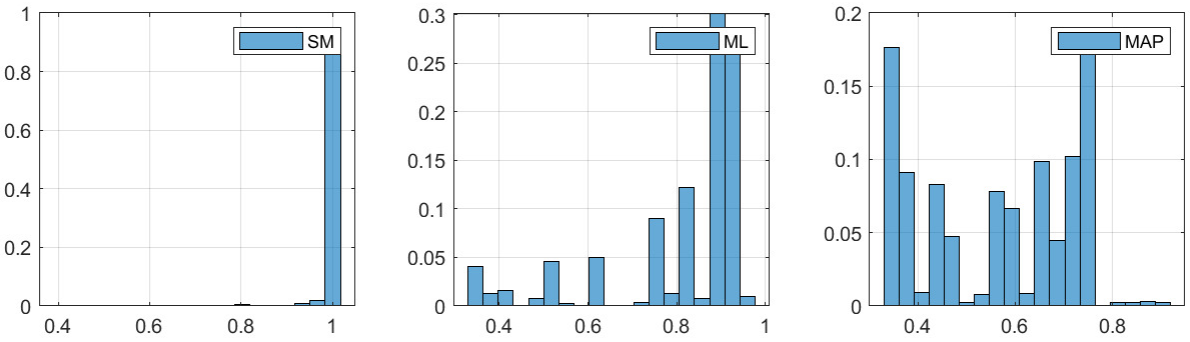}
		\caption{$\lambda_{ML}=5.491\times 10^{-3}$ and $\lambda_{MAP}=5.491\times 10^{-3}$.}
		\label{eff_3_1}
	\end{subfigure}
	\begin{subfigure}[b]{0.45\textwidth}
		\centering
		\includegraphics[scale=0.397]{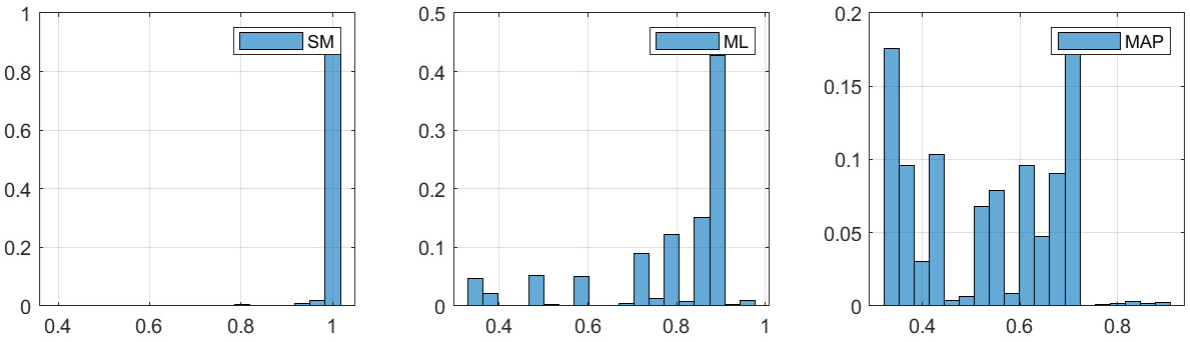}
		\caption{$\lambda_{ML}=6.991\times 10^{-3}$ and $\lambda_{MAP}=6.991\times 10^{-3}$.}
		\label{eff_3_3}
	\end{subfigure}
	\begin{subfigure}[b]{0.45\textwidth}
		\centering
		\includegraphics[scale=0.397]{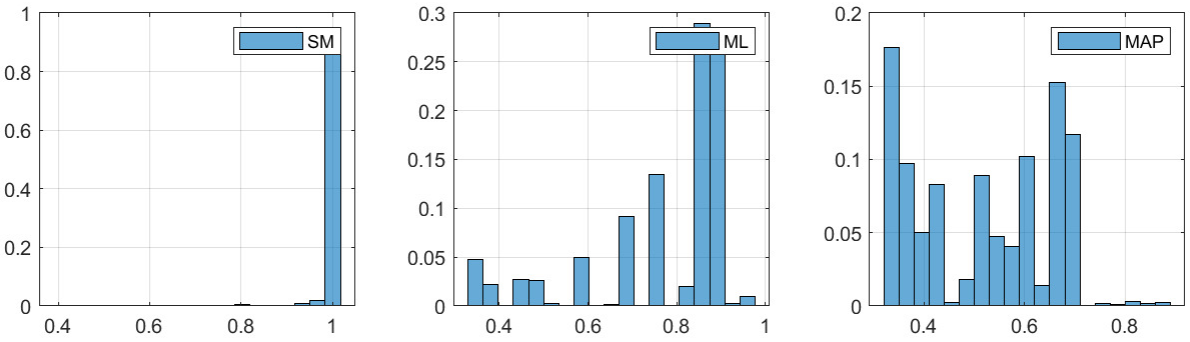}
		\caption{$\lambda_{ML}=7.991\times 10^{-3}$ and $\lambda_{MAP}=7.991\times 10^{-3}$.}
		\label{eff_3_5}
	\end{subfigure}
	\begin{subfigure}[b]{0.45\textwidth}
		\centering
		\includegraphics[scale=0.397]{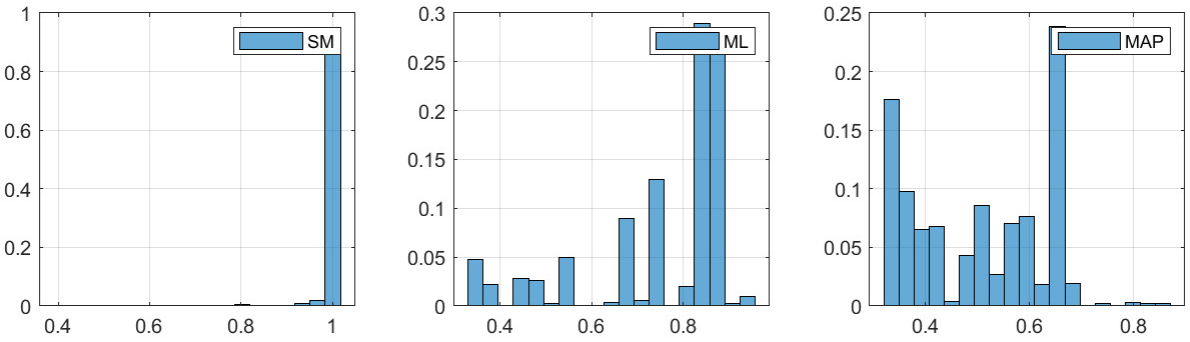}
		\caption{$\lambda_{ML}8.991\times 10^{-3}$ and $\lambda_{MAP}=8.991\times 10^{-3}$.}
		\label{eff_3_7}
	\end{subfigure}
	\begin{subfigure}[b]{0.45\textwidth}
		\centering
		\includegraphics[scale=0.397]{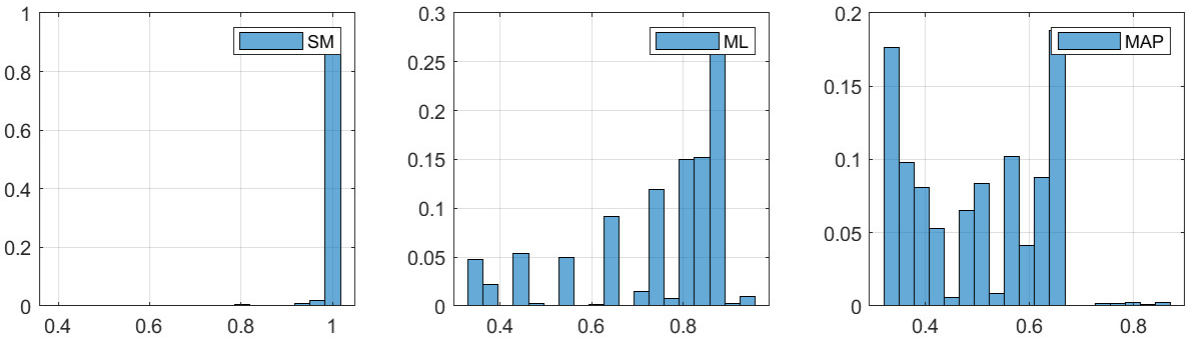}
		\caption{$\lambda_{ML}=9.591\times 10^{-3}$ and $\lambda_{MAP}=9.591\times 10^{-3}$.}
		\label{eff_3_9}
	\end{subfigure}
	\caption{Further results, in terms of the prediction scores (RGB modality), showing the influence of different values of $\lambda$ on the \textit{ML} (center) and the \textit{MAP} (right side). The results using the \textit{SM} layer, in the left-hand side, serves as baseline for comparison. }
	\label{eff_3}
\end{figure}

Additionally to the results presented above, we have implemented the proposed methodology on another state-of-the-art network, the  EfficientNetB1. The performance achieved by the EfficientNetB1 to classify RGB images is a F-score of $98.67\%$ using the Softmax layer (as baseline). The result achieved through the \textit{ML} layer is equivalent to the baseline \ie, F-score = $98.67\%$, while using the \textit{MAP} layer the network achieved $98.66\%$ (almost the same). By keeping $nbins=19$ for both cases, we have performed several runs by changing the values of $\lambda$, and the resulting F-score stabilized around $99.66\%$ \ie, very close to the F-score provided by the Softmax layer (baseline). A way to choose the best values for nbins and $\lambda$ could be, for instance, by reducing the values of the scores of the objects classified as false positives without degrading the results of the true positives, as illustrated by figures \ref{eff_1}, \ref{eff_2}, and \ref{eff_3}, where the distributions in each row were obtained through a given value for the $\lambda$ parameter, considering classifications from the unseen dataset. Note that as the value of $\lambda$ increases, the distributions tend to move away from the extreme values ($0.0$ and $1.0$). 
 \noindent

 \noindent

\noindent
%\appendices
%Appendixes, if needed, appear before the acknowledgment.

%\section*{Acknowledgment}

\bibliographystyle{IEEEtran}
\bibliography{refs}

\begin{thebibliography}{10}
\providecommand{\url}[1]{#1}
\csname url@samestyle\endcsname
\providecommand{\newblock}{\relax}
\providecommand{\bibinfo}[2]{#2}
\providecommand{\BIBentrySTDinterwordspacing}{\spaceskip=0pt\relax}
\providecommand{\BIBentryALTinterwordstretchfactor}{4}
\providecommand{\BIBentryALTinterwordspacing}{\spaceskip=\fontdimen2\font plus
\BIBentryALTinterwordstretchfactor\fontdimen3\font minus
  \fontdimen4\font\relax}
\providecommand{\BIBforeignlanguage}[2]{{%
\expandafter\ifx\csname l@#1\endcsname\relax
\typeout{** WARNING: IEEEtran.bst: No hyphenation pattern has been}%
\typeout{** loaded for the language `#1'. Using the pattern for}%
\typeout{** the default language instead.}%
\else
\language=\csname l@#1\endcsname
\fi
#2}}
\providecommand{\BIBdecl}{\relax}
\BIBdecl

\bibitem{Patel1}
C.~Patel, D.~Bhatt, U.~Sharma, R.~Patel, S.~Pandya, K.~Modi, N.~Cholli,
  A.~Patel, U.~Bhatt, M.~A. Khan, S.~Majumdar, M.~Zuhair, K.~Patel, S.~A. Shah,
  and H.~Ghayvat, ``Dbgc: Dimension-based generic convolution block for object
  recognition,'' \emph{Sensors}, vol.~22, no.~5, 2022.

\bibitem{Bhatt}
D.~Bhatt, C.~Patel, H.~Talsania, J.~Patel, R.~Vaghela, S.~Pandya, K.~Modi, and
  H.~Ghayvat, ``Cnn variants for computer vision: History, architecture,
  application, challenges and future scope,'' \emph{Electronics}, vol.~10,
  no.~20, 2021.

\bibitem{Janai2017}
J.~Janai, F.~Güney, A.~Behl, and A.~Geiger, ``Computer vision for autonomous
  vehicles: Problems, datasets and state of the art,'' \emph{Foundations and
  Trends in Computer Graphics and Vision}, vol.~12, no. 1–3, pp. 1--308,
  2020.

\bibitem{Shaoshan2017}
S.~Liu, L.~Li, J.~Tang, S.~Wu, and J.-L. Gaudiot, ``Creating autonomous vehicle
  systems,'' \emph{Synthesis Lectures on Computer Science}, vol.~6, no.~1, pp.
  i--186, 2017.

\bibitem{Patel2}
C.~I. Patel, S.~Garg, T.~Zaveri, and A.~Banerjee, ``Top-down and bottom-up cues
  based moving object detection for varied background video sequences,''
  \emph{Advances in Multimedia, Hindawi Publishing Corporation}, vol. 2014,
  2014.

\bibitem{hen}
T.~{Hehn}, J.~F.~P. {Kooij}, and D.~M. {Gavrila}, ``Fast and compact image
  segmentation using instance stixels,'' \emph{IEEE Transactions on Intelligent
  Vehicles}, pp. 1--1, 2021.

\bibitem{zwang}
Z.~{Wang}, D.~{Feng}, Y.~{Zhou}, L.~{Rosenbaum}, F.~{Timm}, K.~{Dietmayer},
  M.~{Tomizuka}, and W.~{Zhan}, ``Inferring spatial uncertainty in object
  detection,'' in \emph{EEE/RSJ International Conference on Intelligent Robots
  and Systems}, 2020, pp. 5792--5799.

\bibitem{pcai}
P.~{Cai}, Y.~{Sun}, H.~{Wang}, and M.~{Liu}, ``{VTGNet}: A vision-based
  trajectory generation network for autonomous vehicles in urban
  environments,'' \emph{IEEE Transactions on Intelligent Vehicles}, pp. 1--1,
  2020.

\bibitem{Schutera}
M.~{Schutera}, M.~{Hussein}, J.~{Abhau}, R.~{Mikut}, and M.~{Reischl},
  ``Night-to-day: Online image-to-image translation for object detection within
  autonomous driving by night,'' \emph{IEEE Transactions on Intelligent
  Vehicles}, pp. 1--1, 2020.

\bibitem{hpan}
H.~{Pan}, Z.~{Wang}, W.~{Zhan}, and M.~{Tomizuka}, ``Towards better performance
  and more explainable uncertainty for 3d object detection of autonomous
  vehicles,'' in \emph{IEEE 23rd International Conference on Intelligent
  Transportation Systems}, 2020, pp. 1--7.

\bibitem{cai}
X.~{Cai}, M.~{Giallorenzo}, and K.~{Sarabandi}, ``Machine learning-based target
  classification for {MMW} radar in autonomous driving,'' \emph{IEEE
  Transactions on Intelligent Vehicles}, pp. 1--1, 2021.

\bibitem{zhouli}
C.~{Zhou}, Y.~{Liu}, P.~{Lasang}, and Q.~{Sun}, ``Vehicle detection and
  disparity estimation using blended stereo images,'' \emph{IEEE Transactions
  on Intelligent Vehicles}, pp. 1--1, 2021.

\bibitem{nie}
J.~{Nie}, J.~{Yan}, H.~{Yin}, L.~{Ren}, and Q.~{Meng}, ``A multimodality fusion
  deep neural network and safety test strategy for intelligent vehicles,''
  \emph{IEEE Transactions on Intelligent Vehicles}, pp. 1--1, 2020.

\bibitem{dfeng3}
D.~{Feng}, C.~{Haase-Schütz}, L.~{Rosenbaum}, H.~{Hertlein}, C.~{Gläser},
  F.~{Timm}, W.~{Wiesbeck}, and K.~{Dietmayer}, ``Deep multi-modal object
  detection and semantic segmentation for autonomous driving: Datasets,
  methods, and challenges,'' \emph{IEEE Transactions on Intelligent
  Transportation Systems}, vol.~22, no.~3, pp. 1341--1360, 2021.

\bibitem{CLI}
C.~{Li}, W.~{Xia}, Y.~{Yan}, B.~{Luo}, and J.~{Tang}, ``Segmenting objects in
  day and night: Edge-conditioned cnn for thermal image semantic
  segmentation,'' \emph{IEEE Transactions on Neural Networks and Learning
  Systems}, pp. 1--14, 2020.

\bibitem{zzuo}
Z.~{Zuo}, X.~{Yang}, Z.~{Li}, Y.~{Wang}, Q.~{Han}, L.~{Wang}, and X.~{Luo},
  ``Mpc-based cooperative control strategy of path planning and trajectory
  tracking for intelligent vehicles,'' \emph{IEEE Transactions on Intelligent
  Vehicles}, pp. 1--1, 2020.

\bibitem{Santos}
M.~M.~D. {Santos}, J.~E. {Hoffmann}, H.~G. {Tosso}, A.~W. {Malik}, A.~U.
  {Rahman}, and J.~F. {Justo}, ``Real-time adaptive object localization and
  tracking for autonomous vehicles,'' \emph{IEEE Transactions on Intelligent
  Vehicles}, pp. 1--1, 2020.

\bibitem{Su_2018_ECCV}
D.~Su, H.~Zhang, H.~Chen, J.~Yi, P.-Y. Chen, and Y.~Gao, ``Is robustness the
  cost of accuracy? {A} comprehensive study on the robustness of 18 deep image
  classification models,'' in \emph{European Conference on Computer Vision},
  2018.

\bibitem{Sensoy2018}
M.~Sensoy, L.~Kaplan, and M.~Kandemir, ``Evidential deep learning to quantify
  classification uncertainty,'' in \emph{Advances in Neural Information
  Processing Systems 31}, 2018, pp. 3179--3189.

\bibitem{raudys2003reducing}
{\v{S}}.~Raudys, R.~Somorjai, and R.~Baumgartner, ``Reducing the overconfidence
  of base classifiers when combining their decisions,'' in \emph{Multiple
  Classifier Systems}, 2003, pp. 65--73.

\bibitem{bulatov2015reducing}
K.~B. Bulatov and D.~V. Polevoy, ``Reducing overconfidence in neural networks
  by dynamic variation of recognizer relevance,'' in \emph{Proceedings 29th
  European Conference on Modelling and Simulation}, 2015, pp. 488--491.

\bibitem{kristiadi2020being}
A.~Kristiadi, M.~Hein, and P.~Hennig, ``Being bayesian, even just a bit, fixes
  overconfidence in {R}e{LU} networks,'' in \emph{Proceedings of the 37th
  International Conference on Machine Learning}, vol. 119, 2020, pp.
  5436--5446.

\bibitem{thulasidasan2019mixup}
S.~Thulasidasan, G.~Chennupati, J.~A. Bilmes, T.~Bhattacharya, and S.~Michalak,
  ``On mixup training: Improved calibration and predictive uncertainty for deep
  neural networks,'' in \emph{Advances in Neural Information Processing Systems
  32}, 2019, pp. 13\,888--13\,899.

\bibitem{gupta1}
C.~Gupta, A.~Podkopaev, and A.~Ramdas, ``Distribution-free binary
  classification: prediction sets, confidence intervals and calibration,'' in
  \emph{Advances in Neural Information Processing Systems}, 2020, pp.
  3711--3723.

\bibitem{gupta2}
C.~Gupta and A.~K. Ramdas, ``Top-label calibration and multiclass-to-binary
  reductions,'' in \emph{Proceedings of the International Conference on
  Learning Representations}, 2022.

\bibitem{oncalibration}
C.~Guo, G.~Pleiss, Y.~Sun, and K.~Q. Weinberger, ``On calibration of modern
  neural networks,'' in \emph{Proceedings of the 34th International Conference
  on Machine Learning}, vol.~70, 2017, pp. 1321--1330.

\bibitem{Abdar}
M.~Abdar, F.~Pourpanah, S.~Hussain, D.~Rezazadegan, L.~Liu, M.~Ghavamzadeh,
  P.~Fieguth, X.~Cao, A.~Khosravi, U.~R. Acharya, V.~Makarenkov, and
  S.~Nahavandi, ``A review of uncertainty quantification in deep learning:
  Techniques, applications and challenges,'' \emph{Information Fusion},
  vol.~76, pp. 243--297, 2021.

\bibitem{Mena}
J.~Mena, O.~Pujol, and J.~Vitri\`{a}, ``A survey on uncertainty estimation in
  deep learning classification systems from a bayesian perspective,'' \emph{ACM
  Comput. Surv.}, vol.~54, no.~9, 2021.

\bibitem{Bianca}
B.~Zadrozny and C.~Elkan, ``Obtaining calibrated probability estimates from
  decision trees and naive bayesian classifiers,'' in \emph{Proceedings of the
  Eighteenth International Conference on Machine Learning}.\hskip 1em plus
  0.5em minus 0.4em\relax Morgan Kaufmann Publishers Inc., 2001, p. 609–616.

\bibitem{Naeini}
M.~P. Naeini, G.~F. Cooper, and M.~Hauskrecht, ``Binary classifier calibration:
  Non-parametric approach,'' \emph{arXiv preprint arXiv:1401.3390}, 2014.

\bibitem{GabrielPereyra}
G.~Pereyra, G.~Tucker, J.~Chorowski, L.~Kaiser, and G.~E. Hinton,
  ``Regularizing neural networks by penalizing confident output
  distributions,'' \emph{CoRR, arXiv}, vol. abs/1701.06548, 2017.

\bibitem{posch}
K.~{Posch} and J.~{Pilz}, ``Correlated parameters to accurately measure
  uncertainty in deep neural networks,'' \emph{IEEE Transactions on Neural
  Networks and Learning Systems}, vol.~32, no.~3, pp. 1037--1051, 2021.

\bibitem{zouyu2019}
Y.~{Zou}, Z.~{Yu}, X.~{Liu}, B.~V. K.~V. {Kumar}, and J.~{Wang}, ``Confidence
  regularized self-training,'' in \emph{IEEE International Conference on
  Computer Vision}, 2019, pp. 5981--5990.

\bibitem{Gawlikowski}
J.~Gawlikowski, C.~R.~N. Tassi, M.~Ali, J.~Lee, M.~Humt, J.~Feng, A.~M. Kruspe,
  R.~Triebel, P.~Jung, R.~Roscher, M.~Shahzad, W.~Yang, R.~Bamler, and X.~X.
  Zhu, ``A survey of uncertainty in deep neural networks,'' \emph{CoRR, arXiv},
  vol. abs/2107.03342, 2021.

\bibitem{MartinICCV}
M.~Martin, A.~Roitberg, M.~Haurilet, M.~Horne, S.~Reiss, M.~Voit, and
  R.~Stiefelhagen, ``Drive$\&$act: A multi-modal dataset for fine-grained
  driver behavior recognition in autonomous vehicles,'' in \emph{International
  Conference on Computational Vision}, 2019.

\bibitem{geiger2012}
A.~Geiger, P.~Lenz, and R.~Urtasun, ``Are we ready for autonomous driving? the
  {KITTI} vision benchmark suite,'' in \emph{IEEE Conference on Computer Vision
  and Pattern Recognition}, 2012, pp. 3354--3361.

\bibitem{melotti_icarsc}
G.~{Melotti}, C.~{Premebida}, and N.~{Gonçalves}, ``Multimodal deep-learning
  for object recognition combining camera and {LIDAR} data,'' in \emph{IEEE
  International Conference on Autonomous Robot Systems and Competitions}, 2020,
  pp. 177--182.

\bibitem{yolov420}
A.~Bochkovskiy, C.-Y. Wang, and H.-Y.~M. Liao, ``Yolov4: Optimal speed and
  accuracy of object detection,'' \emph{CoRR, arXiv}, vol. abs/2004.10934,
  2020.

\bibitem{gledson_eccv}
G.~Melotti, C.~Premebida, J.~J. Bird, D.~R. Faria, and N.~Gonçalves,
  ``Probabilistic object classification using {CNN} {ML}-{MAP} layers,'' in
  \emph{Workshop on Perception for Autonomous Driving, European Conference on
  Computer Vision}, 2020.

\bibitem{GledsonMelotti}
G.~Melotti, W.~Lu, D.~Zhao, A.~Asvadi, N.~Gon{\c{c}}alves, and C.~Premebida,
  ``Probabilistic approach for road-users detection,'' \emph{CoRR arxiv}, vol.
  abs/2112.01360, 2021.

\bibitem{shridhar2019}
K.~Shridhar, F.~Laumann, and M.~Liwicki, ``A comprehensive guide to bayesian
  convolutional neural network with variational inference,'' \emph{CoRR,
  arXiv}, vol. abs/1901.02731, 2019.

\bibitem{Graves2011}
A.~Graves, ``Practical variational inference for neural networks,'' in
  \emph{Advances in Neural Information Processing Systems 24}, 2011, pp.
  2348--2356.

\bibitem{Balaji2017}
B.~Lakshminarayanan, A.~Pritzel, and C.~Blundell, ``Simple and scalable
  predictive uncertainty estimation using deep ensembles,'' in \emph{Advances
  in Neural Information Processing Systems 30}, 2017, pp. 6402--6413.

\bibitem{Kendall2017}
A.~Kendall and Y.~Gal, ``What uncertainties do we need in bayesian deep
  learning for computer vision?'' in \emph{Advances in Neural Information
  Processing Systems 30}, 2017, pp. 5574--5584.

\bibitem{Yarin2017}
R.~McAllister, Y.~Gal, A.~Kendall, M.~van~der Wilk, A.~Shah, R.~Cipolla, and
  A.~Weller, ``Concrete problems for autonomous vehicle safety: Advantages of
  bayesian deep learning,'' in \emph{Proceedings of the Twenty-Sixth
  International Joint Conference on Artificial Intelligence}, 2017, pp.
  4745--4753.

\bibitem{Feng2019}
D.~{Feng}, L.~{Rosenbaum}, F.~{Timm}, and K.~{Dietmayer}, ``Leveraging
  heteroscedastic aleatoric uncertainties for robust real-time lidar {3D}
  object detection,'' in \emph{IEEE Intelligent Vehicles Symposium}, 2019, pp.
  1280--1287.

\bibitem{Feng2018}
D.~{Feng}, L.~{Rosenbaum}, and K.~{Dietmayer}, ``Towards safe autonomous
  driving: Capture uncertainty in the deep neural network for lidar {3D}
  vehicle detection,'' in \emph{IEEE 21st International Conference on
  Intelligent Transportation Systems}, 2018, pp. 3266--3273.

\bibitem{yazo}
Y.~Gal and Z.~Ghahramani, ``Dropout as a bayesian approximation: Representing
  model uncertainty in deep learning,'' in \emph{Proceedings of The 33rd
  International Conference on Machine Learning}, vol.~48, 2016, pp. 1050--1059.

\bibitem{jia}
X.~{Jia}, J.~{Yang}, R.~{Liu}, X.~{Wang}, S.~D. {Cotofana}, and W.~{Zhao},
  ``Efficient computation reduction in bayesian neural networks through feature
  decomposition and memorization,'' \emph{IEEE Transactions on Neural Networks
  and Learning Systems}, pp. 1--10, 2020.

\bibitem{AndrewY}
A.~Y. Ng, ``Feature selection, {L1} vs. {L2} regularization, and rotational
  invariance,'' in \emph{Proceedings of the twenty-first international
  conference on Machine learning}, 2004.

\bibitem{lukasik20a}
M.~Lukasik, S.~Bhojanapalli, A.~Menon, and S.~Kumar, ``Does label smoothing
  mitigate label noise?'' in \emph{PMLR Proceedings of the 37th International
  Conference on Machine Learning}, vol. 119, 2020, pp. 6448--6458.

\bibitem{geoffreyhinton}
G.~Hinton, O.~Vinyals, and J.~Dean, ``Distilling the knowledge in a neural
  network,'' in \emph{NIPS Deep Learning and Representation Learning Workshop},
  2015.

\bibitem{corbi}
C.~Corbi\`{e}re, N.~THOME, A.~Bar-Hen, M.~Cord, and P.~P\'{e}rez, ``Addressing
  failure prediction by learning model confidence,'' in \emph{Advances in
  Neural Information Processing Systems}, vol.~32, 2019.

\bibitem{LeaConf}
T.~DeVries and G.~W. Taylor, ``Learning confidence for out-of-distribution
  detection in neural networks,'' \emph{CoRR, arXiv}, vol. abs/1802.04865,
  2018.

\bibitem{BatchNormalization}
S.~Ioffe and C.~Szegedy, ``Batch normalization: Accelerating deep network
  training by reducing internal covariate shift,'' in \emph{Proceedings of the
  32nd International Conference on Machine Learning}, vol.~37, 2015, pp.
  448--456.

\bibitem{drop}
G.~E. Hinton, N.~Srivastava, A.~Krizhevsky, I.~Sutskever, and R.~Salakhutdinov,
  ``Improving neural networks by preventing co-adaptation of feature
  detectors,'' \emph{CoRR, arXiv}, vol. abs/1207.0580, 2012.

\bibitem{Srivastava}
N.~Srivastava, G.~Hinton, A.~Krizhevsky, I.~Sutskever, and R.~Salakhutdinov,
  ``Dropout: A simple way to prevent neural networks from overfitting,''
  \emph{Journal of Machine Learning Research}, vol.~15, pp. 1929--1958, 06
  2014.

\bibitem{DropConnect}
L.~Wan, M.~Zeiler, S.~Zhang, Y.~L. Cun, and R.~Fergus, ``Regularization of
  neural networks using dropconnect,'' in \emph{Proceedings of the 30th
  International Conference on Machine Learning}, vol.~28, no.~3, 2013, pp.
  1058--1066.

\bibitem{isotonicregression}
B.~Zadrozny and C.~Elkan, ``Transforming classifier scores into accurate
  multiclass probability estimates,'' in \emph{Proceedings of the Eighth ACM
  SIGKDD International Conference on Knowledge Discovery and Data Mining},
  2002, pp. 694--–699.

\bibitem{plattscaling}
J.~C. Platt, ``Probabilistic outputs for support vector machines and
  comparisons to regularized likelihood methods,'' in \emph{Advances Large
  Margin Classifiers}, 2000, pp. 61--74.

\bibitem{betacalibration}
M.~Kull, T.~Silva~Filho, and P.~Flach, ``Beta calibration: a well-founded and
  easily implemented improvement on logistic calibration for binary
  classifiers,'' in \emph{20th AISTATS.}, 2017, pp. 623--631.

\bibitem{mix}
J.~Zhang, B.~Kailkhura, and T.~Han, ``Mix-n-match: Ensemble and compositional
  methods for uncertainty calibration in deep learning,'' in
  \emph{International Conference on Machine Learning (ICML)}, 2020.

\bibitem{Chen_2019_CVPR}
Q.~Chen, W.~Zhang, J.~Yu, and J.~Fan, ``Embedding complementary deep networks
  for image classification,'' in \emph{IEEE/CVF Conference on Computer Vision
  and Pattern Recognition}, 2019, pp. 9230--9239.

\bibitem{LIANG2019}
Y.~Liang, H.~Huang, Z.~Cai, Z.~Hao, and K.~C. Tan, ``Deep infrared pedestrian
  classification based on automatic image matting,'' \emph{Applied Soft
  Computing}, vol.~77, pp. 484 -- 496, 2019.

\bibitem{Papoulis}
A.~Papoulis and U.~Pillai, \emph{\BIBforeignlanguage{English (US)}{Probability,
  random variables and stochastic processes}}, 4th~ed.\hskip 1em plus 0.5em
  minus 0.4em\relax McGraw-Hill, Nov. 2001.

\bibitem{bishop}
C.~M. Bishop, \emph{Pattern Recognition and Machine Learning}.\hskip 1em plus
  0.5em minus 0.4em\relax Springer, 2006.

\bibitem{ScottDavidW}
D.~W. Scott, \emph{Multivariate density estimation : theory, practice, and
  visualization}, ser. Wiley series in probability and mathematical
  statistics.\hskip 1em plus 0.5em minus 0.4em\relax Wiley, 1992.

\bibitem{AdditiveS}
D.~Valcarce, J.~Parapar, and {\'A}.~Barreiro, ``Additive smoothing for
  relevance-based language modelling of recommender systems,'' in
  \emph{Proceedings of the 4th Spanish Conference on Information Retrieval},
  2016.

\bibitem{SmoTec}
S.~F. Chen and J.~Goodman, ``An empirical study of smoothing techniques for
  language modeling,'' Harvard Computer Science Group Technical Report, Tech.
  Rep., 1998.

\bibitem{Lidstone}
G.~J. Lidstone, ``Note on the general case of the bayes-laplace formula for
  inductive or a posteriori probabilities,'' \emph{Transactions of the Faculty
  of Actuaries}, vol.~8, p. 182–192, 1920.

\bibitem{incv3}
C.~{Szegedy}, V.~{Vanhoucke}, S.~{Ioffe}, J.~{Shlens}, and Z.~{Wojna},
  ``Rethinking the inception architecture for computer vision,'' in \emph{IEEE
  Conference on Computer Vision and Pattern Recognition}, 2016, pp. 2818--2826.

\bibitem{beytscal}
M.~Kull, M.~Perello~Nieto, M.~K\"{a}ngsepp, T.~Silva~Filho, H.~Song, and
  P.~Flach, ``Beyond temperature scaling: Obtaining well-calibrated multi-class
  probabilities with dirichlet calibration,'' in \emph{Advances in Neural
  Information Processing Systems 32}, 2019, pp. 12\,316--12\,326.

\bibitem{Niculescu}
A.~Niculescu-Mizil and R.~Caruana, ``Predicting good probabilities with
  supervised learning,'' in \emph{Proceedings of the 22nd International
  Conference on Machine Learning}, 2005, pp. 625--632.

\bibitem{nuscenes20}
H.~Caesar, V.~Bankiti, A.~H. Lang, S.~Vora, V.~E. Liong, Q.~Xu, A.~Krishnan,
  Y.~Pan, G.~Baldan, and O.~Beijbom, ``nu{S}cenes: A multimodal dataset for
  autonomous driving,'' in \emph{IEEE Conference on Computer Vision and Pattern
  Recognition}, June 2020.

\bibitem{astar3d}
Q.~H. {Pham}, P.~{Sevestre}, R.~S. {Pahwa}, H.~{Zhan}, C.~H. {Pang}, Y.~{Chen},
  A.~{Mustafa}, V.~{Chandrasekhar}, and J.~{Lin}, ``{A*3D} dataset: Towards
  autonomous driving in challenging environments,'' in \emph{IEEE International
  Conference on Robotics and Automation}, 2020, pp. 2267--2273.

\bibitem{Song_2019_CVPR}
X.~Song, P.~Wang, D.~Zhou, R.~Zhu, C.~Guan, Y.~Dai, H.~Su, H.~Li, and R.~Yang,
  ``{ApolloCar3D}: A large {3D} car instance understanding benchmark for
  autonomous driving,'' in \emph{IEEE/CVF Conference on Computer Vision and
  Pattern Recognition}, 2019, pp. 5447--5457.

\bibitem{Yang_2019_CVPR}
G.~Yang, X.~Song, C.~Huang, Z.~Deng, J.~Shi, and B.~Zhou, ``{DrivingStereo}: A
  large-scale dataset for stereo matching in autonomous driving scenarios,'' in
  \emph{IEEE/CVF Conference on Computer Vision and Pattern Recognition}, 2019,
  pp. 899--908.

\bibitem{H3D2019}
A.~{Patil}, S.~{Malla}, H.~{Gang}, and Y.~{Chen}, ``The {H3D} dataset for
  full-surround {3D} multi-object detection and tracking in crowded urban
  scenes,'' in \emph{IEEE International Conference on Robotics and Automation},
  2019, pp. 9552--9557.

\bibitem{Mapillary2017}
G.~{Neuhold}, T.~{Ollmann}, S.~R. {Bulò}, and P.~{Kontschieder}, ``The
  mapillary vistas dataset for semantic understanding of street scenes,'' in
  \emph{IEEE International Conference on Computer Vision}, 2017, pp.
  5000--5009.

\bibitem{lyftlevel5}
L.~Vincent, P.~Ondruska, A.~Jain, S.~Omari, and V.~Shet, ``Tutorial:
  Perception, prediction, and large scale data collection for autonomous
  cars,'' in \emph{IEEE Conference on Computer Vision and Pattern Recognition},
  2019.

\bibitem{lyft2019}
R.~Kesten, M.~Usman, J.~Houston, T.~Pandya, K.~Nadhamuni, A.~Ferreira, M.~Yuan,
  B.~Low, A.~Jain, P.~Ondruska, S.~Omari, S.~Shah, A.~Kulkarni, A.~Kazakova,
  C.~Tao, L.~Platinsky, W.~Jiang, and V.~Shet, ``Lyft level 5 perception
  dataset 2020,'' \url{https://level5.lyft.com/dataset/}, 2019.

\bibitem{me1}
G.~Melotti, C.~Premebida, N.~M. M. d.~S. Goncalves, U.~J.~C. Nunes, and D.~R.
  Faria, ``Multimodal cnn pedestrian classification: A study on combining lidar
  and camera data,'' in \emph{21st International Conference on Intelligent
  Transportation Systems (ITSC)}, 2018, pp. 3138--3143.

\bibitem{me2}
G.~Melotti, A.~Asvadi, and C.~Premebida, ``Cnn-lidar pedestrian classification:
  combining range and reflectance data,'' in \emph{IEEE International
  Conference on Vehicular Electronics and Safety (ICVES)}, 2018, pp. 1--6.

\bibitem{cp_high}
C.~Premebida, L.~Garrote, A.~Asvadi, A.~P. Ribeiro, and U.~Nunes,
  ``High-resolution lidar-based depth mapping using bilateral filter,'' in
  \emph{2016 IEEE 19th International Conference on Intelligent Transportation
  Systems (ITSC)}, 2016, pp. 2469--2474.

\end{thebibliography}

\begin{IEEEbiography}[{\includegraphics[width=1in,height=2.0in,clip,keepaspectratio]{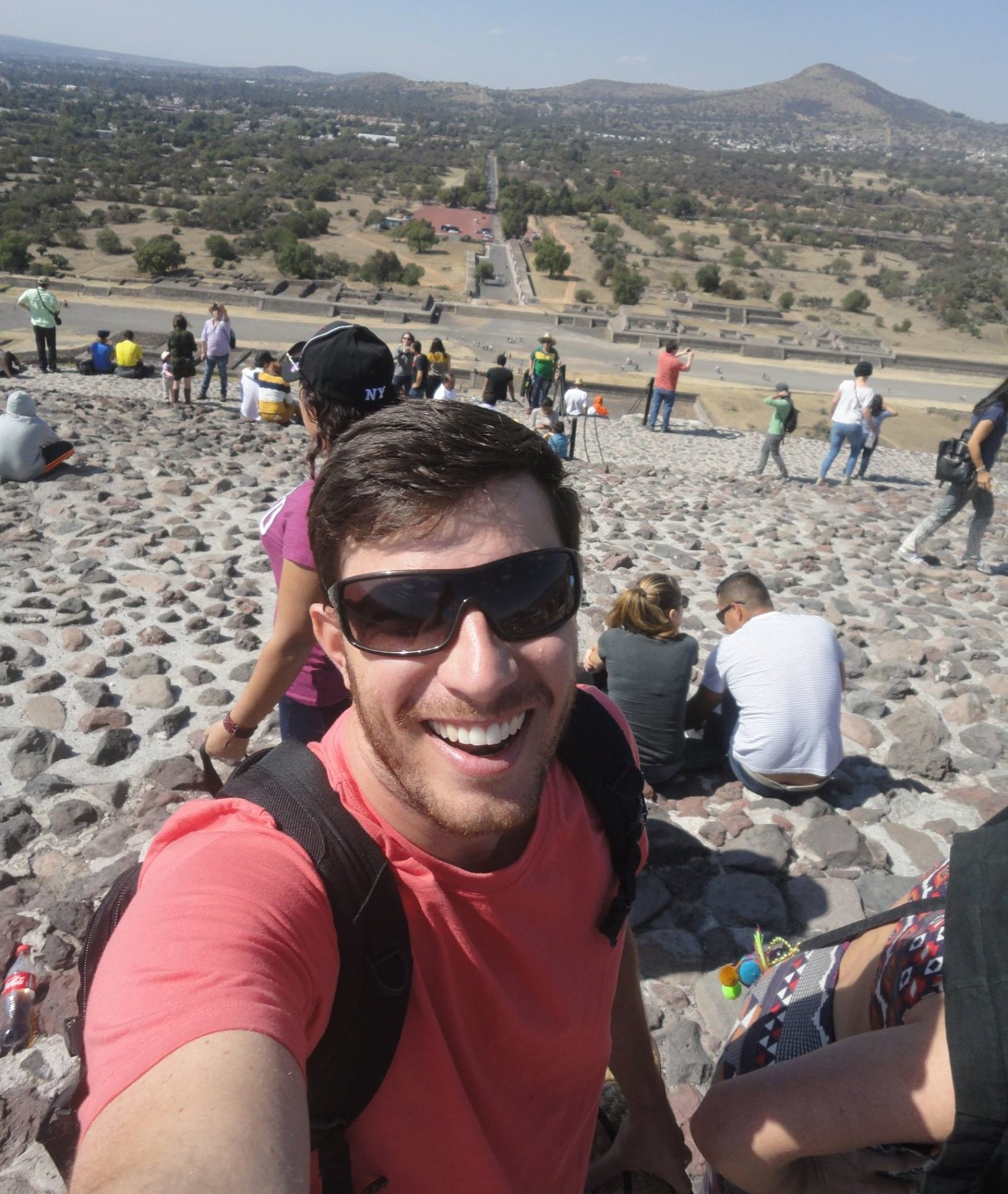}}]{Gledson Melotti}
	received a Bachelor's degree in Electrical Engineering from the Federal University of Sao Joao del-Rei-MG-Brazil, in 2006. In 2009 he received a master's degree in Electrical Engineering from the Federal University of Minas Gerais-MG-Brazil. He is currently pursuing a Ph.D. degree from the Department of Electrical and Computer Engineering at University of Coimbra-Portugal. His research interests are confidence calibration, deep learning, point clouds, and sensor fusion strategies applied to autonomous driving perception.
\end{IEEEbiography}

\begin{IEEEbiography}[{\includegraphics[width=1in,height=2.0in,clip,keepaspectratio]{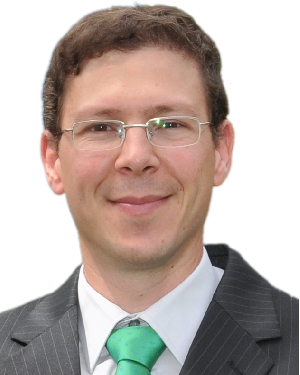}}]{Cristiano Premebida} is Assistant Professor in the department of electrical and computer engineering at the University of Coimbra, Portugal, where he is a member of the Institute of Systems and Robotics (ISR-UC). Between September 2018 and December 2019, he worked as lecturer in autonomous vehicles in the AAE department at the Loughborough University, UK. His main research interests are robotic perception, machine learning, Bayesian inference, autonomous vehicles, autonomous robots, agricultural robotics, and sensor fusion. He works on multimodal and multisensory perception for robotics and autonomous systems applications, developing calibration strategies and probability-prediction approaches to increase robustness of deep models.
\end{IEEEbiography}

\begin{IEEEbiography}[{\includegraphics[width=1in,height=1.25in,clip,keepaspectratio]{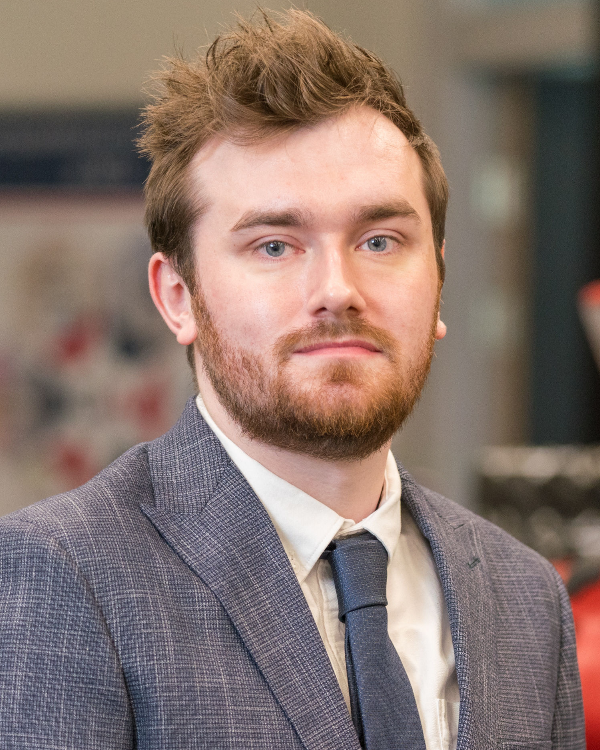}}]{Jordan J. Bird} is a Research Fellow with the Computational Intelligence and Applications Research Group (CIA) within the Department of Computer Science at Nottingham Trent University, UK. Before that, he studied for a PhD in Human-Robot Interaction at Aston University. Jordan’s research interests include Artificial Intelligence (AI), Human-Robot Interaction (HRI), Machine Learning (ML), Deep Learning, Transfer Learning, and Data Augmentation.
\end{IEEEbiography}
%\vfill
\begin{IEEEbiography}[{\includegraphics[width=1in,height=1.25in,clip,keepaspectratio]{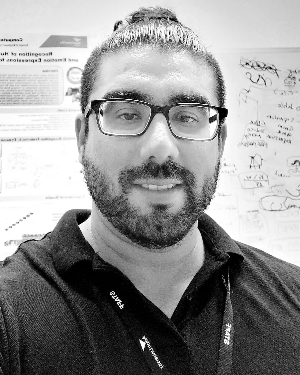}}]{Diego R. Faria} is a Reader (Associate Professor) in Robotics and Adaptive Systems. He is with the School of Physics, Engineering and Computer Science, University of Hertfordshire, Hatfield, UK. Previously, from 07-2016 to 02-2022 he was a Lecturer and Senior Lecturer (from 2019) at Aston University, UK. Currently (2019-2022) he is the coordinator of the EU CHIST-ERA InDex project (Robot In-hand Dexterous manipulation by extracting data from human manipulation of objects to improve robotic autonomy and dexterity) funded by EPSRC UK. Dr Faria is also PI (2020-2022) of projects with industry (KTP-Innovate UK scheme) related to perception and autonomous systems applied to autonomous vehicles. He received his Ph.D. degree in electrical and computer engineering from the University of Coimbra, Portugal in 2014. He holds an M.Sc. degree in computer science from the Federal University of Parana, Brazil, in 2005. In 2001, he earned a bachelor’s degree in informatics technology (data computing \& information) and has finished a computer science specialization in 2002 at Londrina State University, Brazil. From 2014 to 2016 he was a postdoctoral fellow at the Institute of Systems and Robotics, University of Coimbra where he collaborated on different projects funded by EU commission and the Portuguese government in areas of Robot Grasping, Artificial Perception, Cognitive Robotics (HRI), Assistive Technology and Applied Machine Learning, including Bayesian Inference. 
\end{IEEEbiography}

\begin{IEEEbiography}[{\includegraphics[width=1in,height=1.25in,clip,keepaspectratio]{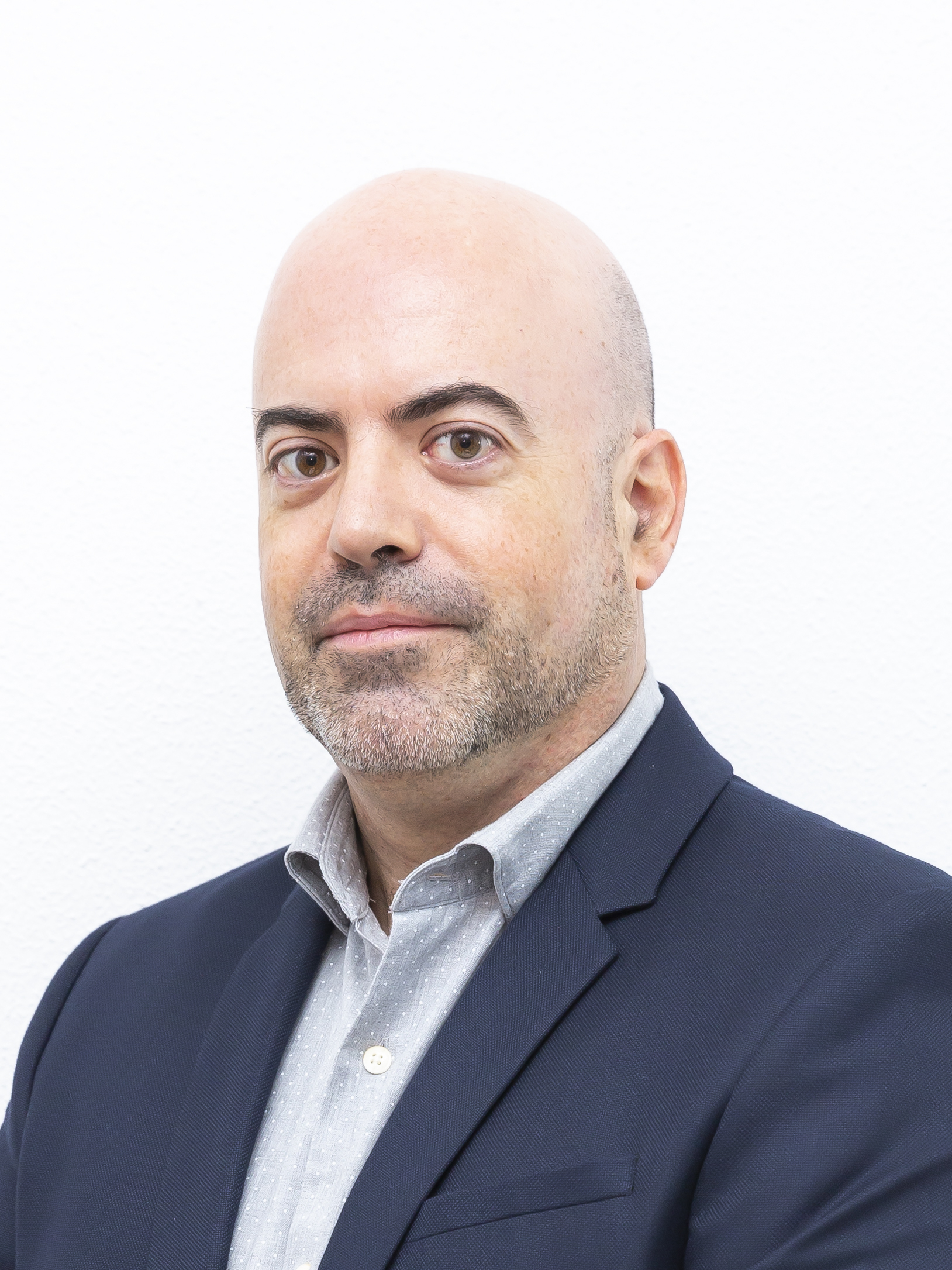}}]{Nuno Gon\c calves}
	received the Ph.D. degree in computer vision from the University of Coimbra, Portugal, in 2008. Since 2008, he has been a tenured Assistant Professor with the Department of Electrical and Computers Engineering, Faculty of Sciences and Technologies, University of Coimbra. He is currently a Senior Researcher with the Institute of Systems and Robotics, University of Coimbra. He has been recently coordinating several projects centered on the technology transfer to the industry. In 2018, he joined the Portuguese Mint and Official Printing Office (INCM), where he coordinates innovation projects in areas, such as facial recognition, graphical security, information systems, and robotics. He has been working in the design and introduction of new products as result of the innovation projects. He is the author of several articles and communications in high-impact journals and international conferences. His scientific career has been mainly developed in the fields of computer vision, visual information security, and robotics, but also in computer graphics.
\end{IEEEbiography}

\end{document}